\documentclass{article}

\usepackage[preprint]{nips}

\usepackage[utf8]{inputenc} 
\usepackage[T1]{fontenc}    
\usepackage{hyperref}       
\usepackage{url}            
\usepackage{booktabs}       
\usepackage{amsfonts}       
\usepackage{nicefrac}       
\usepackage{microtype}      
\usepackage{xcolor}         
\usepackage{multirow}
\usepackage{graphicx}
\usepackage{caption}
\usepackage{subcaption}
\usepackage{algorithm}
\usepackage{algpseudocode}
\usepackage{booktabs}
\usepackage{wrapfig}
\usepackage{amsmath}

\title{Multimodal ELBO with Diffusion Decoders
}

\author{%
  Daniel Wesego and Pedram Rooshenas \\ 
   Department of Computer Science\\
  University of Illinois Chicago\\
  \texttt{\{dweseg2,pedram\}@uic.edu} 
}

\usepackage{tcolorbox}

\newcommand{\mbf}[1]{\mathbf{#1}}

\newcommand{\mbfm}[1]{\mathbf{#1}_{1:M}}

\begin{document}

\maketitle
\begin{abstract}
Multimodal variational autoencoders have demonstrated their ability to learn the relationships between different modalities by mapping them into a latent representation. Their design and capacity to perform any-to-any conditional and unconditional generation make them appealing. However, different variants of multimodal VAEs often suffer from generating low-quality output, particularly when complex modalities such as images are involved. In addition to that, they frequently exhibit low coherence among the generated modalities when sampling from the joint distribution. To address these limitations, we propose a new variant of the multimodal VAE ELBO that incorporates a better decoder using a diffusion generative model. The diffusion decoder enables the model to learn complex modalities and generate high-quality outputs. The multimodal model can also seamlessly integrate with a standard feed-forward decoder for different types of modality, facilitating end-to-end training and inference. Furthermore, we introduce an auxiliary score-based model to enhance the unconditional generation capabilities of our proposed approach. This approach addresses the limitations imposed by conventional multimodal VAEs and opens up new possibilities to improve multimodal generation tasks. Our model provides state-of-the-art results compared to other multimodal VAEs in different datasets with higher coherence and superior quality in the generated modalities.

\end{abstract}

\section{Introduction}

Deep learning has revolutionized many fields through its ability to extract and learn patterns that are otherwise difficult through other means \citep{lecun2015deep, SCHMIDHUBER201585}. One area that has seen immense progress is generative models, where neural networks learn to generate samples that are similar to the training distribution. Variational autoencoders (VAEs) have emerged as one powerful framework for this task \citep{vae, rezende14}. VAEs were initially built to process a single modality, such as an image. However, in the real world, data often comes from multiple modalities that are interconnected and expressed in different formats \citep{Baltrusaitis}. Multimodal learning aims to build models that can process and relate information from various modalities, such as vision, language, and other modalities together \citep{mul_AndrewNG}. Incorporating this multimodal information can have a huge impact on the performance of the model in tasks that require understanding from multiple perspectives \citep{JMLR:v15:srivastava14b}.

Multimodal variational autoencoders extend the VAE framework to learn joint representations across modalities \citep{mul_AndrewNG, suzuki2016joint, poe}. By capturing the correlations between modalities, these models can learn robust and informative representations. These learned multimodal representations can be used in different ways, including transfer learning in downstream tasks \citep{mvtcae, microvideo_transfer}, and they also provide a framework to perform unconditional and conditional generation across the different modalities \citep{mmvae, wesego}. 

\begin{figure*}[!t]
    \centering
    \begin{minipage}{.48\textwidth}
        \centering
        \includegraphics[width=\textwidth]{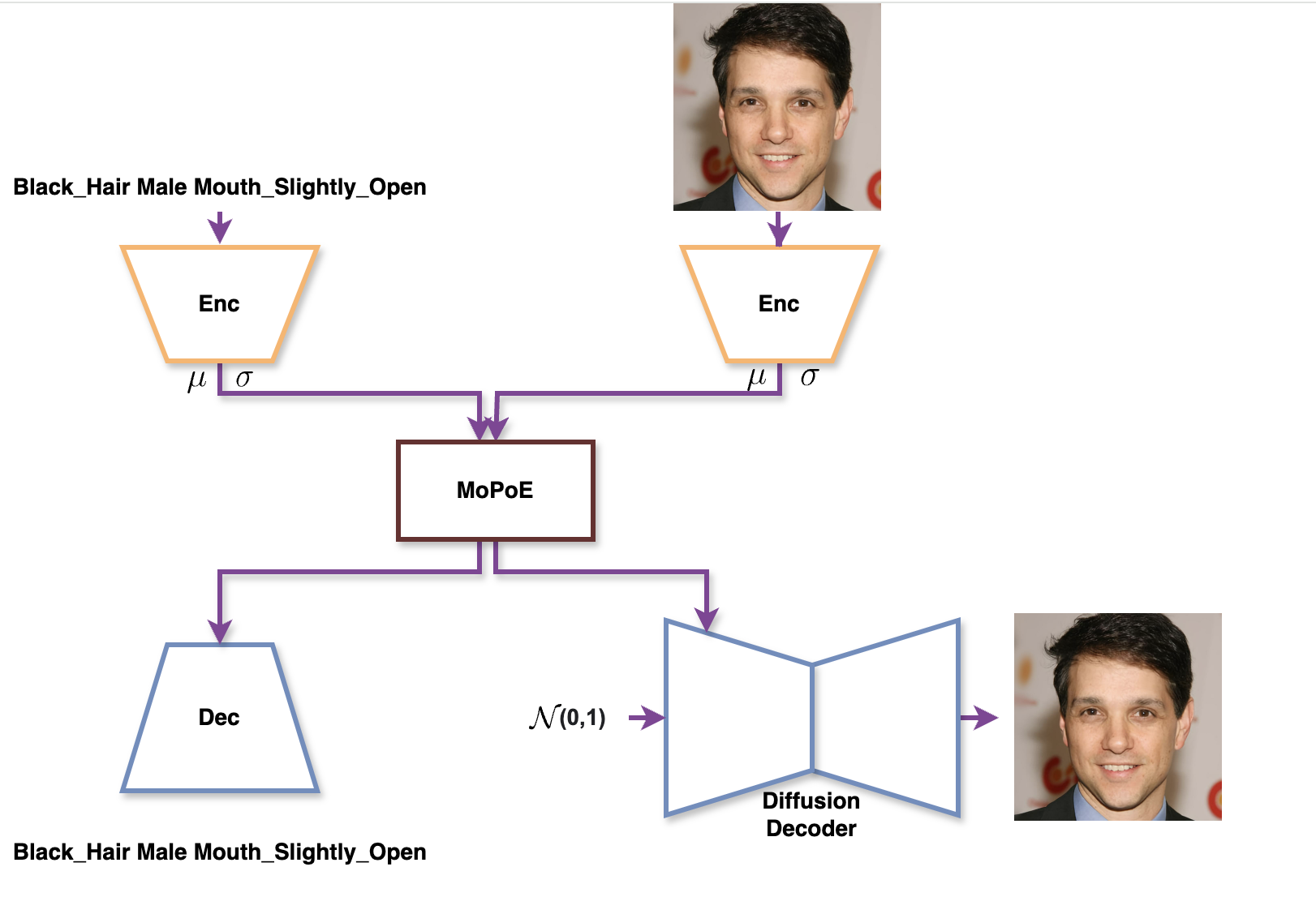}
    \caption{Architecture of our model where modality-specific encoders map the input data from different modalities into latent space, which are then fused using a mixture of PoE and passed to the respective decoders.}
    \label{fig:design1}
    \end{minipage}%
    \hspace{0.02\textwidth}
    \begin{minipage}{.48\textwidth}
        \centering
        \includegraphics[width=\textwidth]{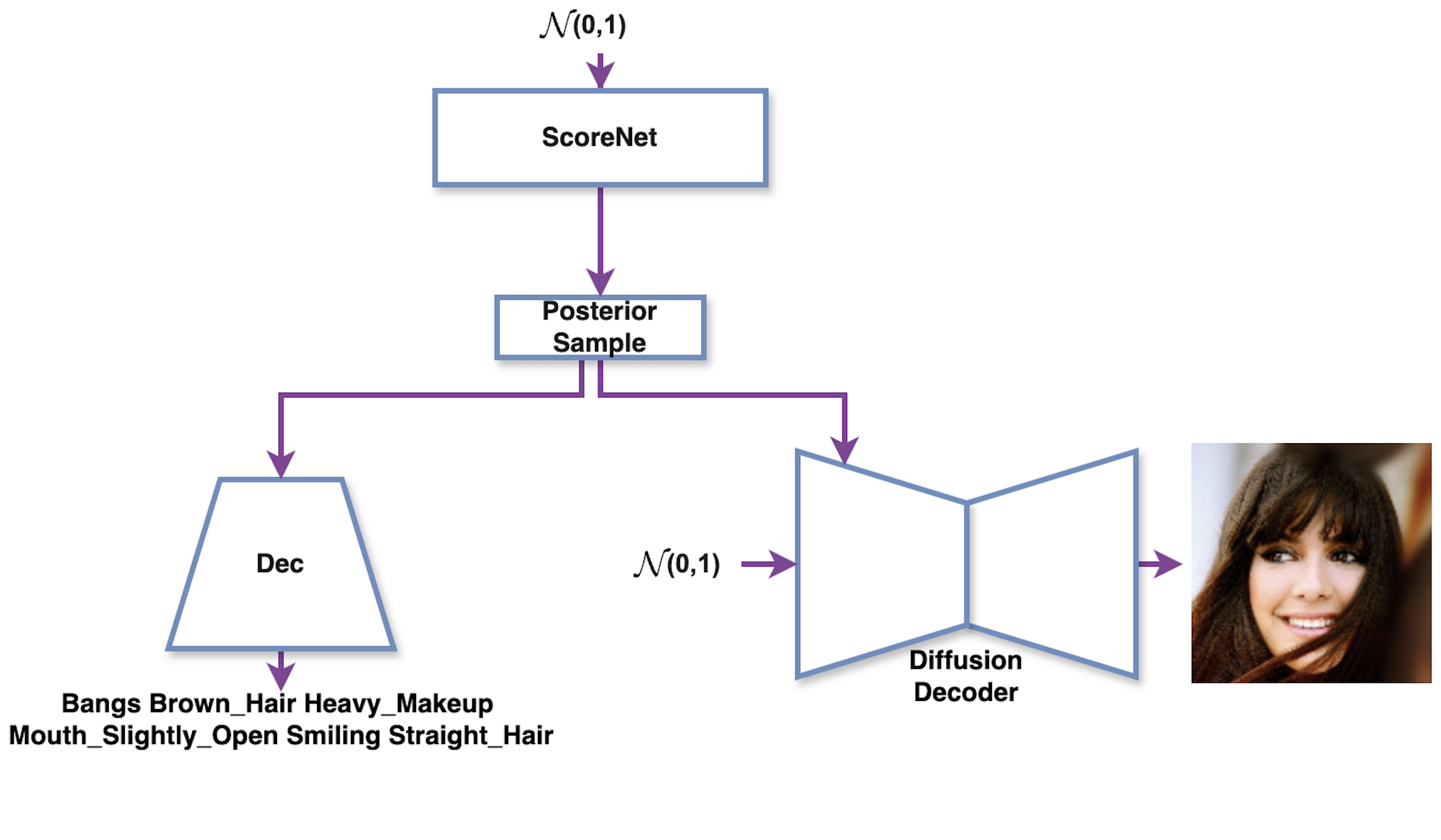}
        \vspace{0.02\textwidth}
    \caption{Unconditional sampling technique that leverages a score-based model to transform a Gaussian noise distribution into the PoE posterior distribution of all modalities.}
    \label{fig:design2}
    \end{minipage}%
\end{figure*}


Despite their promising potential, multimodal VAEs continue to face significant challenges in effectively integrating diverse modalities, generating high-quality samples, and scaling to high-dimensional data \citep{daunhawer2022on, wesego}. Various prior works have attempted to address these issues by proposing alternative ELBO objectives, yet the core problems still persist in multimodal VAEs \citep{mopoe, jsd}. Due to these ongoing limitations, most multimodal studies have been constrained to low-dimensional datasets such as MNIST or PolyMNIST \citep{mnist, jsd}. Additionally, \citet{daunhawer2022on} highlights the generative discrepancies in mixture-based models and raises concerns about the practicality of these models in real-world applications.


On the other hand, diffusion models have emerged as a powerful class of generative models, gaining substantial attention in recent years for their ability to produce high-quality samples \cite{diff_ho_etal, dhariwal}. These models operate by gradually transforming a data distribution into a noise distribution, with the goal of learning to reverse this process. The forward process is a Markovian sequence that incrementally adds noise to the data distribution over several timesteps, while the reverse process, typically modeled by a neural network, learns to remove the noise and reconstruct the data \citep{diff_ho_etal}. A notable advantage of diffusion models is their capacity to generate exceptionally high-quality samples, achieving state-of-the-art performance in various generative tasks, including image synthesis and inpainting \citep{photorealistic, lugmayr2022repaint}. \citet{preechakul2021diffusion} introduced diffusion autoencoder, where the encoder resembles that of a VAE, but the decoder is modeled using a diffusion process. The key difference between a diffusion decoder and a traditional feed-forward decoder used in VAEs is that the diffusion decoder conditions the generation process on the latent representation rather than directly feeding it as input. This approach enables the diffusion autoencoder to learn meaningful representations as well as generate high-quality samples.

In this study, we introduce a novel multimodal generative model that combines the strengths of diffusion autoencoders and multimodal VAEs to learn a unified joint representation across multiple modalities. Traditional multimodal VAEs often face limitations in generating high-quality outputs from their joint representations, which may not capture all the details of all modalities ~\cite{palumbo2022mmvae,mvtcae}. These challenges are further exacerbated by the low-performance decoders typically used in standard VAEs, resulting in suboptimal model performance. To address these issues, we propose the use of a flexible diffusion decoder, which allows the compressed joint representation to effectively generate high-quality outputs for complex modalities while leveraging feedforward decoders for simpler modalities.


To achieve this, we propose a multimodal ELBO that is integrated with diffusion decoders for modalities where feed-forward decoders have shown limitations, such as images, ensuring high-quality sample generation and improved representation capacity. Conversely, for modalities where standard decoders have proven effective, such as text or sparse outputs, we retain standard VAE decoders. This hybrid approach allows our model to harness the strengths of each generative model class while effectively addressing their respective limitations.

Our model achieves better coherence in conditional generation among the generated modalities while generating high-quality samples. We use a mixture of product of experts (MoPoE) to fuse the representations from different modalities \citep{mopoe}. This approach enables our model to capture the correlations and dependencies between modalities. For unconditional generation tasks, where the goal is to generate samples from all modalities simultaneously without conditioning on any specific modality, we propose the use of an auxiliary model. This auxiliary model transforms a Gaussian noise distribution into the approximate posterior product of experts distribution. By leveraging this technique, our model gains increased flexibility and can perform well in both conditional and unconditional generation settings.

Figure \ref{fig:design1} illustrates the general design and architecture of our proposed model. The modality-specific encoders map the input data from different modalities into latent representations, which are then fused using the MoPoE mechanism. The fused representation is then passed to the respective decoders, which can be either diffusion decoder or a standard one, depending on the modality. Figure \ref{fig:design2} shows how to perform unconditional sampling where the score-based model is employed to provide the necessary input distributions to the decoders.

Our proposed model demonstrates superior performance compared to previous variants of multimodal variational autoencoder models. Our key contributions can be summarized as follows:

\begin{itemize}
    \item We introduce a novel generative multimodal autoencoder architecture that seamlessly integrates diffusion and standard decoders. This hybrid design allows our model to generate high-quality samples across modalities without compromising the coherence and consistency among the generated modalities. In doing so, our architecture strikes an optimal balance between sample quality and cross-modal coherence.
    \item To enhance the model's performance in unconditional generation settings, we propose to use a sampling technique that employs an auxiliary score-based model that transforms a Gaussian noise distribution into the approximate distribution of the PoE posterior to improve the model's ability to generate coherent and diverse samples across all modalities in an unconditional manner for multimodal VAEs.
    \item Our approach can be scaled to high-dimensional data modalities with good generative quality compared to previous approaches, which opens a new research direction toward multimodal VAEs.
\end{itemize}

Through these contributions, our model addresses the critical limitations of previous multimodal VAE approaches while retaining their desirable properties and extending their capabilities to handle high-dimensional data modalities effectively. Our approach paves the way for more powerful and flexible multimodal generative models capable of generating high-quality samples with enhanced coherence.

\section{Related Works}

Multimodal learning has emerged as a promising research direction in the deep learning field that has started from early works such as \citet{mul_AndrewNG, yao_learning_factorized_multimodal}. \citet{mul_AndrewNG} demonstrated the potential of multimodal deep learning models in leveraging information from multiple modalities to improve performance on multiple tasks by learning combined features. As variational autoencoders (VAEs) gained traction for simultaneously learning representations and generation models within individual modalities, researchers soon explored extending the VAE framework to handle multiple modalities simultaneously, giving rise to multiple variants of multimodal VAEs.

\citet{suzuki2016joint} first presented a model that is trained jointly and can accept multiple modalities. They propose training additional encoders for each modality and minimizing the KL loss between the joint ones to handle missing modalities. \citet{poe} proposed a product of experts approach to encode the modalities jointly. They posterior is constructed as $q_\phi(\mathbf{z|x}) = p(\mathbf{z} )\prod_i^M q_{\phi}(\mathbf{ z}|\mathbf {x_i})$. This approach allowed different formulations of the ELBO in future works because the product of experts can be calculated in a closed form, and it allows the generation of missing modalities by simply plugging $1$ in the product for the missing modalities. The model proposed by \citet{poe}, which is called MVAE, generates images comparative to a unimodal VAE but has low coherence amongst the generated modalities. 

\citet{mmvae} proposed a mixture of experts posterior formulation that samples one modality at a time from the set containing all the modalities. Their model shows better coherence but, in general, has degraded quality, especially visible in image modalities. \citet{mopoe} further expanded the subsets in the mixture by proposing a mixture of product of experts posterior. \citet{daunhawer2022on} showed that models that subsample will have a loss in generative quality because of reconstructing all available modalities from a few of them. Other approaches by \citet{prvshared, palumbo2022mmvae, jsd} explored splitting the latent space into modality-specific and shared representations to improve the generative quality of multimodal VAEs.

In the work by \citet{mvtcae}, the authors propose an evidence lower bound (ELBO) objective derived from an information-theoretic perspective that uses a product of experts approach. Their main objective is to learn a representation that minimizes the conditional total correlation between the joint conditional and the factorized conditionals. Their final objective is a convex combination of two terms: one based on the variational information bottleneck (VIB) and the other based on the conditional VIB.

~\citet{diffusevae} proposed a two-stage training in which a unimodal VAE is trained in the first stage, and a diffusion model conditioned on the output of the VAE is trained in the second stage to refine the lower-quality output of a VAE. Similar approaches have also been proposed in multimodal works to refine the outputs of the image modalities of multimodal VAEs by ~\citet{palumbo2024deep, wesego}, but the diffusion model is not directly integrated to the ELBO like our proposed work.

Alternatively, \citet{wesego, e26040320} proposed to use unimodal VAEs to represent each modality and use an additional neural network to fuse the modalities instead of a product/mixture of experts using a two-step training process. They both select a score-based network to join the modalities in the latent space and sample from the joint distribution. \citet{rombach2021highresolution} proposed text-to-image generation that employs a high-quality autoencoder to compress images into latent space, allowing for diffusion processes to occur in this reduced dimensionality. They integrated CLIP models using cross-attention to effectively represent and condition text inputs. Building upon this, ~\citet{xu2023versatile, any-to-any-codi} proposed multi-modal latent diffusion models that can generate any modality from any given condition. They expand \citet{rombach2021highresolution} work to use multiple diffusion flows to achieve this. It's worth noting that these advanced models require training on extremely large-scale datasets, which requires extensive computational resources. 

\citet{preechakul2021diffusion} introduced diffusion autoencoder as a means for diffusion models to generate images and learn representations. They transformed the diffusion model into an autoencoder form where an encoder learns a latent space that will be used in the reconstruction process. ~\citet{hudson2023soda} modified the diffusion autoencoder to reconstruct multiple views in the diffusion decoder. ~\citet{infodiff} proposed a diffusion autoencoder architecture where the latent variable $\mathbf{z}$ is encouraged to learn disentangled meaningful representation by regularizing the objective with mutual information among $\mathbf{x}$ and $\mathbf{z}$.

In this work, we focus on improving multimodal VAEs by proposing an objective that flexibly combines both diffusion autoencoder and standard VAEs. Overall, \citet{daunhawer2022on} illustrated that the current multimodal VAEs are very limited due to their lack of generation capability as the modalities become more complex. This is caused by how the multimodal ELBO objective is constructed and by the limitation in VAEs to generate high-quality images \citep{diffusevae}. Our approach enhances the generative quality of multimodal VAEs while still preserving the fundamental framework and strengths of the multimodal VAE architecture in conditional and unconditional settings.

\section{Methodology}

We first present our method starting with diffusion autoencoders. Subsequently, we introduce the multimodal VAE framework that incorporates diffusion autoencoders. The section concludes by explaining how to perform both conditional and unconditional sampling techniques.

\subsection{Diffusion Autoencoder}
A diffusion model can be formulated as a latent variable model where the marginal distribution is expressed as $p_\theta(\mathbf{x}_0) = \int p_\theta(\mathbf{x}_{0:T}) \mathrm{d}\mathbf{x}_{1:T}$ \citep{diff_ho_etal}. The combined probability distribution $p_\theta(\mathbf{x}_{0:T}) = p(\mathbf{x}_T) \prod_{t=1}^T p_\theta(\mathbf{x}_{t-1} | \mathbf{x}_t)$ is the reverse joint process where $p(\mathbf{x}_T) = \mathcal{N}(\mathbf{0},\mathbf{I})$. One can sample any intermediate time step $t$ in the forward process starting from the data point $\mathbf{x}_0$ using the Gaussian distribution $q(\mathbf{x}_t | \mathbf{x}_0) = \mathcal{N}(\mathbf{x}_t; \sqrt{\bar{\alpha}_t} \mathbf{x}_0, (1 - \bar{\alpha}_t) \mathbf{I})$ to add noise to the data. A neural network is trained to reverse this process where $p_\theta(\mathbf{x}_{t-1} | \mathbf{x}_t) = \mathcal{N}(\mathbf{x}_{t-1}; \mu_\theta(\mathbf{x}_t, t), \Sigma_\theta(\mathbf{x}_t, t))$. \citet{diff_ho_etal} observed that using a simplified objective and predicting the noise is better and easier to implement. The objective is shown in eq.~\ref{eq:diff_simple}.
\begin{align}
\label{eq:diff_simple}
    L(\theta) = \mathbb{E}_{t,\mathbf{x}_0,\boldsymbol{\epsilon}}\left[\left\|\boldsymbol{\epsilon} - \epsilon_\theta(\mathbf{x}_t, t)\right\|^2\right]
\end{align}
where $\mathbf{x_t} = \sqrt{\bar{\alpha}_t} \mathbf{x_0} + \sqrt{1 - \bar{\alpha}_t} \epsilon_t$ and $\bar{\alpha}_t$ are schedule hyperparameters.

A diffusion autoencoder combines the architectural setup of an autoencoder with a diffusion decoder \citep{preechakul2021diffusion}. An encoder $q(\mathbf{z|x})$ projects the data into a latent representation $\mathbf{z}$. This latent representation is fed to the diffusion decoder which generates samples conditioned on $\mathbf{z}$. By training the model this way, the diffusion model can be used as a powerful representation where the encoder learns a meaningful latent representation of the data similar to an autoencoder \citep{hudson2023soda}. The objective for a diffusion autoencoder is very similar to the diffusion objective, except we have additional conditioning from the encoder. It is trained by minimizing $\mathbb{E}_{\mathbf{x_0, \epsilon_t}} [\left\| \epsilon - \epsilon_{\theta}(\mathbf{x}_t, \mathbf{z}, t) \right\|^2]$ where $\mathbf{z}$ comes from the encoder network $q(\mathbf{z|x})$.

In addition, diffusion models can be trained to maximize the log-likelihood of training data by using a specific weight on the score-matching loss term \citep{song2021maximum}. By using this objective, which is upper bounded by the log-likelihood, diffusion models can achieve log-likelihoods that are comparable to the best models. Specifically, with a stochastic differential equation (SDE) of the form $d\mathbf{x} = f(\mathbf{x}, t) dt + g(t) d\mathbf{w}$ that diffuses data to noise with a drift coefficient $f$ and a diffusion coefficient $g$ where $\mathbf{w}$  is a standard Wiener process, we can use a weighting $\lambda(t) = g(t)^2 $ for likelihood weighting. The denoising score-matching loss to train diffusion models for likelihood training is shown in eq. ~\ref{eq:sm_ll}, where additional conditioning term $\mathbf{z}$ is added from the encoder to form the autoencoder setup and a Monte Carlo estimate is used for the expectation with $t$ sampled from uniform distribution, $\mathbf{x}$ from the training data, and $s_\theta()$ denotes the score function \citep{song2021maximum}. 
\begin{align} \label{eq:sm_ll}
   L(\theta) = \, \mathbb{E}_{p(\mathbf{x}) p(t) p(\mathbf{z|x})p_t(\mathbf{x}_t \mid \mathbf{x}, \mathbf{z})} \Bigg[ \frac{1}{2} \lambda(t) \Big\| \nabla_{\mathbf{x}_t} \log p_t(\mathbf{x}_t \mid \mathbf{x}, \mathbf{z}) - s_\theta(\mathbf{x}_t, \mathbf{z}, t) \Big\|_2^2 \Bigg] 
\end{align}

\subsection{Multimodal VAE with Diffusion Decoder}
Assuming that we have M modalities, the observed data is represented as $\mbfm{x} = {\mathbf{x}^{(1)}, \mathbf{x}^{(2)}, \ldots, \mathbf{x}^{(M)}}$. Mixture-based multimodal models encode the posterior distribution using a mixture of experts of Gaussian distributions $q_\phi^S(\mathbf{z} | \mathbf{x}) = \sum_{A \in \mathcal{S}} \omega_A q_\phi(\mathbf{z} | \mathbf{x}^A)$ where $0 \le \omega_A \leq 1$ and $\mathcal{S}$ is a selected set of modalities formed from the powerset of the combination of modalities. Using a mixture of experts allows easy inference when some of the modalities are not observed, and training can be performed by sampling one component from the mixture at a time. ~\citet{daunhawer2022on} highlighted that mixture-based multimodal models face an inherent limitation compared to unimodal VAEs due to the sub-sampling of modalities in the encoder. In these models, a restricted subset of modalities is used to encode information for all modalities, requiring the decoders to reconstruct all modalities from this limited input. This challenge is particularly evident in multimodal VAEs, where the decoder often struggles with the insufficient information provided by the encoder. However, incorporating a more powerful decoder can help address this issue effectively.
For any subset $\mathcal{S}$ of modalities, a multimodal evidence lower bound (ELBO) can be derived for the log-likelihood of the multimodal data ~\cite{daunhawer2022on}: 
\begin{align} \label{eq:m_elbo_l}
     \log p_\theta(\mbfm x) \ge \sum_{A \in \mathcal{S}} \omega_A \Bigg\{ 
    \mathbb{E}_{q_{\phi}(\mathbf{z} \mid \mathbf{x}^A)} 
    \left[\log p_\theta(\mbfm{x} \mid \mathbf{z}) \right] - D_\text{KL} \big(q_{\phi}(\mathbf{z} \mid \mathbf{x}^A) \,\Vert\, p(\mathbf{z})\big) \Bigg\}.
\end{align}

The full derivation is given in the Appendix ~\ref{appx:elbo_proof} for reference. \citet{daunhawer2022on} demonstrates that this bound generalizes various multimodal VAE objectives, such as MoPoE~\cite{mopoe}, MMVAE~\cite{mmvae}, and MVAE~\cite{poe}.



Assuming that the modalities are conditionally independent given the latent representation $\mathbf{z}$ (i.e. $\mbf x^i \perp \mbf x^{\setminus i} \mid \mbf z$), the ELBO can be written the following:
\begin{align} \label{eq:m_elbo_l2}
    \log p(\mbfm x) \ge \sum_{A \in \mathcal{S}} \omega_A \Bigg\{ 
    \mathbb{E}_{q_{\phi}(\mathbf{z} \mid \mathbf{x}_A)} 
    \left[ \sum_{i=1}^M \log p_\theta(\mathbf{x}^i \mid \mathbf{z}) \right] - D_\text{KL} \big(q_{\phi}(\mathbf{z} \mid \mathbf{x}^A) \,\Vert\, p(\mathbf{z})\big) \Bigg\}.
\end{align}

These conditional independencies enable us to support modality-specific decoders -- specifically, feed-forward and diffusion-based decoders. To distinguish between them, we denote the number of modalities with feed-forward decoders as $M_F$ and those with diffusion-based decoders as $M_D$.


\textbf{Proposition 1}: 
The objective defined in eq. ~\ref{eq:m_elbo_l3} is a valid lower bound on the marginal likelihood of the data under the proposed model. 
\begin{align}
\label{eq:m_elbo_l3}
\log &p(\mbfm{x}) \ge \nonumber \sum_{A \in \mathcal{S}} \omega_A \Bigg\{ \mathbb{E}_{q_{\phi}(\mathbf{z}| \mathbf{x}_A)} \Bigg[ \sum_{i=1}^{M_F} \log p_\theta(\mathbf{x}_i| \mathbf{z}) + \nonumber \sum_{j=1}^{M_D} \mathbb{E}_{\mathbf{x}_{jt}} 
 \frac{1}{2} \lambda(t) \bigg\| \nabla_{\mathbf{x}_{jt}} \log p_t(\mathbf{x}_{jt} \mid \mathbf{x}_j, \mathbf{z}) \nonumber \\ 
 & - s_\theta(\mathbf{x}_{jt}, \mathbf{z}, t) \bigg\|_2^2 \Bigg] - D_\text{KL} \big(q_{\phi}(\mathbf{z}|\mathbf{x}_A) \big\| p(\mathbf{z})\big) \Bigg\}.
\end{align}

See Appendix ~\ref{appx:elbo_proof} for the proof.




\subsection{Conditional Sampling}

To perform conditional sampling (i.e. generate missing modalities given some modalities) utilizing our model, we can leverage the factorized nature of the posterior distribution. We first sample from the joint posterior distribution conditioned on the available modalities and then use the sampled latent variable to generate samples using the respective generative model for each modality. The posterior will be the product of the observed modalities $q_\phi(\mathbf{z}|\mathbf{x}_{o}) = \prod_{i \in {o}} q_\phi(\mathbf{z}_i|\mathbf{x}_{i})$ where the subscript $o$ denotes observed modalities. Then, we can generate an unobserved modality ${u}$ from the posterior $q_\phi(\mathbf{z}|\mathbf{x}_{o})$ by feeding it to the appropriate decoder $p(\mathbf{x|z)}$. By following this, we can generate any modality from any other given modality.

\subsection{Unconditional Sampling}

Unconditional generation occurs when no specific modality is provided as input. In this scenario, the goal is to produce samples that exhibit coherence across all modalities without relying on any initial conditions. Previous multimodal VAEs perform unconditional generation by first sampling from a $\mathcal{N}(0,\mathbf{I})$ and then feeding the sample to each decoder. Doing that may produce suboptimal results as there is a gap between the posterior and the prior \citep{ncpvae}. To get better samples, we train an auxiliary score-based diffusion model to sample from the posterior. Figure ~\ref{fig:design2} demonstrates how the score model facilitates unconditional sampling. This auxiliary score model is trained using the diffusion objective but on the latent space $\mathbf{z}$ to generate samples from the posterior. The objective is $\mathbb{E}_{t,\mathbf{z}_0,\boldsymbol{\epsilon}}[\left\|\boldsymbol{\epsilon} - \epsilon_\theta(\mathbf{z}_t, t)\right\|^2$ where $\mathbf{z_0}$ is sampled from $q_\phi(\mathbf{z}|\mathbf{x}) = \prod_{i} q_\phi(\mathbf{z}_i|\mathbf{x}_{i})$ where $\mathbf{x}_i$ consists of each modality in $\mathbf{x}$. The main objective of training this auxiliary model is to get high-quality samples from the posterior instead of using $\mathcal{N}(0,\mathbf{I})$ as the starting sample for unconditional generation. To generate samples unconditionally, we first sample $\mathbf{z}$ from the auxiliary score model and use the respective generative model $p_\theta(\mathbf x_i|\mathbf{z})$ of each modality to get the final samples.

\section{Experiments}
\label{experiments}

We perform experiments on two datasets. The first one is from \citet{WahCUB_200_2011}, which contains images of birds with text captions describing the birds. We use actual images of the birds instead of ResNET features, unlike previous multimodal VAEs that struggle with image data \citep{mmvae}. The second dataset is a high-dimensional CelebAMask-HQ ~\citep{CelebAMask-HQ} dataset that consists of images, masks, and attributes of celebrities expressed in the three modalities. We compare different multimodal VAE baselines consisting of MVTCAE \citep{mvtcae}, MoPoE \citep{mopoe}, and MMVAE+ \citep{palumbo2022mmvae}. MMVAE+ is not included in the CUB dataset experiment as we use pre-trained VAE for the text captions, which will require changing the structure of the text VAE because of the modality-specific and shared representations in MMVAE+. Nonetheless, we added the results from their paper for the text-to-image results, which can be referenced from their work \citep{palumbo2022mmvae}. We also add additional baselines from ~\citet{wesego, e26040320} that use a latent score-based diffusion model to learn the joint latent space discussed in the next sections. We compare these baselines with our proposed model variants Diff-MVAE and Diff-MVAE*. Diff-MVAE* is trained using the same objective as eq.~\ref{eq:m_elbo_l3}, but with $\lambda(t) = 1$. This choice introduces a mismatch between the ELBO and the score-matching objective, resulting in a violation of the ELBO in Eq.~\ref{eq:m_elbo_l3}. However, empirical results indicate that setting $\lambda(t) = 1$ leads to higher-quality generated samples, a finding that has also been reported in previous works~\citep{diff_ho_etal, lsgm}.

\subsection{Caltech Birds (CUB)}


\begin{wrapfigure}{r}{0.5\textwidth}
    \centering
    \begin{subfigure}{1\textwidth}
    \begin{subfigure}{0.48\textwidth}
    \centering
    \begin{subfigure}{0.48\textwidth}
    \centering
     \includegraphics[width=0.5\textwidth]{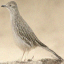}
    \end{subfigure}
    \begin{subfigure}{0.48\textwidth}
    \centering
     \includegraphics[width=0.5\textwidth]{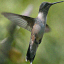}
    \end{subfigure}
     \begin{tcolorbox}[colback=white, boxrule=0.1mm, left=0.5mm, right=0.5mm, boxsep=0.05mm, equal height group=myg1]
        \centering
         \tiny	
         a small sized bird that has multiple tones of grey all over
        \end{tcolorbox}
    \end{subfigure}
    \end{subfigure}
    \caption{Conditional samples given text using Diff-MVAE (left), Diff-MVAE*(right)}
    \label{fig:mddvae_cond_gtext_cub}
\end{wrapfigure}

The CUB dataset consists of two modalities: image and text describing the image (caption). In order to get meaningful text outputs, we initialized the text VAE from the pre-trained weight of ~\citet{li2020_Optimus} that uses a BERT encoder and GPT-2 decoder for all models. The images are resized to 64x64. We used a latent size of $768$ for both the image and text. The encoder architecture for the image modality is similar to that of ~\citet{softvae} and the diffusion decoder uses a UNET architecture. The models were trained for 500 epochs using the Adam optimizer \citep{adam}. The baseline models are trained using different $\beta$ values listed in the appendix, and the best model was selected by using the average of the conditional and unconditional FID. The unconditional auxiliary score model also uses a 1D UNET architecture and accepts input similar to the size of the latent dimension. DDIM sampling was used for 50 steps to generate samples for the image modality for Diff-MVAE* and for unconditional sampling \citep{ddim}. We used Euler–Maruyama sampling for Diff-MVAE. More details about the training and inference are in the Appendix section. 

\begin{wrapfigure}{r}{0.5\textwidth}
    \centering
    \begin{subfigure}[!]{1\textwidth}
    \begin{subfigure}[!]{0.23\textwidth}
        \centering
         \includegraphics[width=0.48\textwidth]{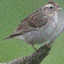}
         \begin{tcolorbox}[colback=white, boxrule=0.1mm, left=0.5mm, right=0.5mm, boxsep=0.05mm, equal height group=myg1]
         \centering
         \tiny
         this bird has wings that are brown and has a white belly
        \end{tcolorbox}
    \end{subfigure}
    \begin{subfigure}[!]{0.23\textwidth}
    \centering
     \includegraphics[width=0.48\textwidth]{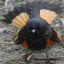}
     \begin{tcolorbox}[colback=white, boxrule=0.1mm, left=0.5mm, right=0.5mm, boxsep=0.05mm, equal height group=myg1]
        \centering
         \tiny	
         a small bird with a long bill and bright orange breast
        \end{tcolorbox}
    \end{subfigure}
    \end{subfigure}
    \caption{Unconditional samples using Diff-MVAE (left), Diff-MVAE*(right)}
    \label{fig:mddvae_unc_cub}
\end{wrapfigure}

We use FID, which is a widely adopted metric for evaluating the quality of images, as the evaluation metric to measure conditionally and unconditionally generated samples  \citep{fid}. In addition to that, we use Clip-Score to measure the similarity of the image-text pairs ~\citep{hessel-etal-2021-clipscore, e26040320}. It's clear that a good multimodal VAE model will have coherent unconditional samples where the image and text represent the same concept expressed in its respective modality. In addition, conditionally generated images should follow the description in the text, and conditionally generated text should follow the given image.

\begin{table}[!h]
\caption{CUB FID Result}
\label{cub_results_fid}
\begin{center}
\begin{tabular}{ccc}
\toprule
{} & {Txt-to-Img} & {Unc} \\
\midrule
Diff-MVAE  & 57.4\tiny{($\pm$0.2)} &  70.4\tiny{($\pm$0.05)}\\
Diff-MVAE*  & \textbf{35.5}\tiny{($\pm$0.5)} & \textbf{37.5}\tiny{($\pm$0.1)} \\
MoPoE  & 290.6\tiny{($\pm$0.5)} & {199.6}\tiny{($\pm$0.75)} \\
MVTCAE & {176.3}\tiny{($\pm$0.05)} & 168.9\tiny{($\pm$0.2)}   \\
\midrule
MLD & 62.6 & 63.4\\
MMVAE+ & 164.9 & - \\
\bottomrule
\end{tabular}
\end{center}
\end{table}

We present the results in Table ~\ref{cub_results_fid} and ~\ref{cub_results_clip}. In addition to the implemented baselines, we add results from previous works of MLD by ~\citet{e26040320} and MMVAE+ on text-to-image FID. The FID scores evaluation demonstrates that our model, Diff-MVAE, outperforms the baselines, generating high-quality images that exhibit greater coherence and alignment with the text modality. In contrast, the baselines struggle to generate meaningful images that are consistent with the textual information, supporting the findings of \citet{daunhawer2022on} regarding the challenges faced by existing multimodal VAEs in difficult tasks. Table ~\ref{cub_results_clip} also shows that Diff-MVAE generates more coherent conditional outputs that follow the given modality. We also added qualitative results from our model, which is the only multimodal VAE model generating good images in this dataset, are shown in figure ~\ref{fig:mddvae_cond_gtext_cub} and ~\ref{fig:mddvae_unc_cub} for conditional and unconditional generation respectively.


\begin{table}[!b]
\caption{CUB Clip Score Result}
\label{cub_results_clip}
\begin{center}
\begin{tabular}{cccc}
\toprule
{} & {Img-to-Txt} & {Txt-to-Img} & {Unc} \\
\midrule
Diff-MVAE  & 27.7\tiny{($\pm$0.01)} & \textbf{28.8}\tiny{($\pm$0.02)} & \textbf{28.23}\tiny{($\pm$0.01)} \\
Diff-MVAE*  & \textbf{28.1}\tiny{($\pm$0.001)} & {28.52}\tiny{($\pm$0.02)} & {28.12}\tiny{($\pm$0.005)}  \\
MoPoE  & 26.8\tiny{($\pm$0.03)} & {22.0}\tiny{($\pm$0.03)} & 24.8\tiny{($\pm$0.05)} \\
MVTCAE & {27.1}\tiny{($\pm$0.03)} & 26.6\tiny{($\pm$0.004)} & 28.08\tiny{($\pm$0.02)}  \\
\bottomrule
\end{tabular}
\end{center}
\end{table}

\subsection{CelebAMask-HQ}

The CelebAMask-HQ dataset provides multimodal representations of visual characteristics related to human faces. In this dataset, the images, masks, and attributes can be viewed as distinct yet complementary modalities that capture various aspects of an individual's appearance. We resize the images and masks to a resolution of 128 by 128 pixels. We combine multiple masks of parts of the face into a single 0/1 mask similar to ~\citet{wesego}. Specifically, all the provided masks in the CelebAMask-HQ dataset, except for the skin mask, are drawn on top of each other, resulting in a single composite mask. This composite mask encodes the presence or absence of various facial features. For the attribute modality, we follow the preprocessing approach described by \citet{poe, wesego}. Out of the 40 existing attributes in the CelebAMask-HQ dataset, we selectively choose 18 attributes that are most relevant and informative for representing the visual characteristics of a person's face. We used a latent size of $256$. Similar to the previous dataset, the encoder and decoder architecture for the image modality are inspired by ~\citet{softvae} except for the diffusion decoder, which uses a UNET architecture. The unconditional auxiliary score-based model also works on the $256$ dimension latent size resized to 16x16 utilizing a UNET architecture. We used a similar sampling strategy of the previous dataset for Diff-MVAE variants. More details about the experimental setup are located in the appendix section. The baselines here are MoPoE, MVTCAE, MMVAE+ plus SBM-VAE-C ~\citep{wesego} and MLD ~\citep{e26040320}, which use unimodal variational autoencoders (VAEs) or autoencoders (AEs) for each modality and fuse them using a score-based model for learning the joint distribution. We also add MoPoE*, which is trained with the same hyperparameters as Diff-MVAE but without diffusion decoders.

\begin{figure}[!h]
    \centering
    \begin{subfigure}[!]{1\textwidth}
        \begin{subfigure}[!]{0.24\textwidth}
            \centering
             \includegraphics[width=1\textwidth]{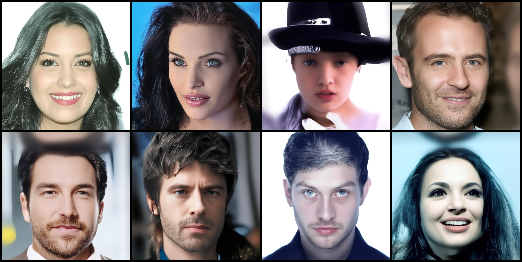}
             \caption{Diff-MVAE}
        \end{subfigure}
        \begin{subfigure}[!]{0.24\textwidth}
            \centering
             \includegraphics[width=1\textwidth]{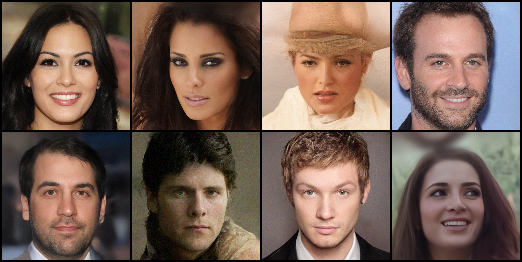}
             \caption{Diff-MVAE*}
        \end{subfigure}
        \begin{subfigure}[!]{0.24\textwidth}
            \centering
             \includegraphics[width=1\textwidth]{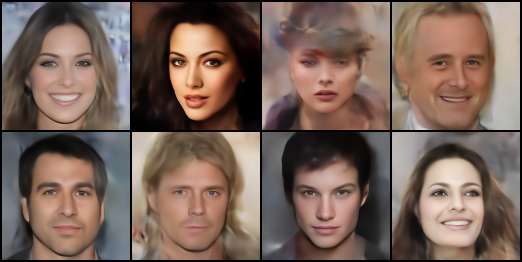}
             \caption{MVTCAE}
        \end{subfigure}
        \begin{subfigure}[!]{0.24\textwidth}
            \centering
             \includegraphics[width=1\textwidth]{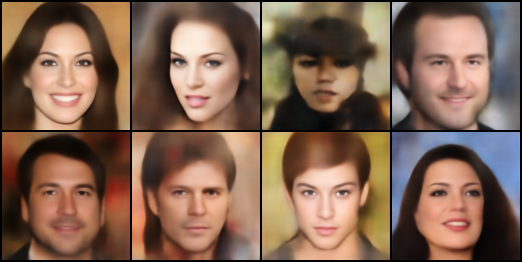}
             \caption{MMVAE+}
        \end{subfigure}
    \end{subfigure}
    \caption{Conditional generated images for different models given the same masks and attribute for all the models}
    \label{fig:mddvae_cond}
\end{figure}

\begin{table*}[!]
\caption{CelebAMask-HQ Result}
\label{celeba_result}
\small{
\begin{center}
\resizebox{\textwidth}{!}{%
\begin{tabular}{ccc|cccc|cc}
\toprule
\multicolumn{1}{c}{} &\multicolumn{2}{c}{Attribute} &\multicolumn{4}{c}{Image} &\multicolumn{2}{c}{Mask}  \\
\midrule
 GIVEN &\multicolumn{1}{c}{Both} &\multicolumn{1}{c}{Img} & {Both} & {Mask} & {Attr} & {Unc} &\multicolumn{1}{c}{Both} &\multicolumn{1}{c}{Img} \\
\midrule
{} & {F1} & {F1} & {FID} & {FID} & {FID} &{FID} &  {F1} & {F1} \\
\midrule

Diff-MVAE  & \textbf{0.76}\tiny{($\pm$0.001)} & {0.75}\tiny{($\pm$0.001)} & 42.3\tiny{($\pm$0.2)} & {41.9}\tiny{($\pm$0.06)} & 
{45.1}\tiny{($\pm$0.45)} & {43.5}\tiny{($\pm$0.2)} &  \textbf{0.90}\tiny{($\pm$0.001)} & {0.90}\tiny{($\pm$0.001)}  \\
Diff-MVAE*  & \textbf{0.76}\tiny{($\pm$0.001)} & \textbf{0.76}\tiny{($\pm$0.001)} & \textbf{28.3}\tiny{($\pm$0.02)} & \textbf{28.5}\tiny{($\pm$0.1)} & \textbf{32.3}\tiny{($\pm$0.35)} & \textbf{28.4}\tiny{($\pm$0.3)} &  \textbf{0.90}\tiny{($\pm$0.001)} & {0.90}\tiny{($\pm$0.001)}  \\
MoPoE*  & 0.74\tiny{($\pm$0.002)} & {0.75}\tiny{($\pm$0.002)} & 104.8\tiny{($\pm$0.31)} & 105.8\tiny{($\pm$0.18)} & 182.3\tiny{($\pm$0.29)} & 139.8\tiny{($\pm$0.65)}  & \textbf{0.90}\tiny{($\pm$0.001)} & {0.90}\tiny{($\pm$0.001)} \\
MoPoE  & 0.68\tiny{($\pm$0.002)} & {0.71}\tiny{($\pm$0.004)} & 114.9\tiny{($\pm$0.32)} & 101.1\tiny{($\pm$0.16)} & 186.7\tiny{($\pm$0.28)} & 164.8\tiny{($\pm$0.62)}  & 0.85\tiny{($\pm$0.002)} & \textbf{0.92}\tiny{($\pm$0.001)} \\
MVTCAE & {0.71}\tiny{($\pm$0.001)} & 0.69\tiny{($\pm$0.004)} & 94\tiny{($\pm$0.45)} & 84.2\tiny{($\pm$0.32)} & 87.2\tiny{($\pm$0.08} & 162.2\tiny{($\pm$1.08)} & {0.89}\tiny{($\pm$0.001)} & 0.89\tiny{($\pm$0.003)} \\
MMVAE+ & 0.64\tiny{($\pm$0.003)} & 0.61\tiny{($\pm$0.002)} & 133\tiny{($\pm$14.28)} & 97.3\tiny{($\pm$0.04)} & 153\tiny{($\pm$0.49)} & 103.7\tiny{($\pm$0.61)} & 0.82\tiny{($\pm$0.03)} & 0.89\tiny{($\pm$0.002)} \\
SBM-VAE-C  & 0.69\tiny{($\pm$0.005)} & 0.66\tiny{($\pm$0.001)} & 82.4\tiny{($\pm$0.1)} & {81.7}\tiny{($\pm$0.29)} & {76.3}\tiny{($\pm$0.7)} &  {79.1}\tiny{($\pm$±0.3)} & 0.84\tiny{($\pm$0.02)} & 0.84\tiny{($\pm$0.001)}  \\
MLD  & 0.71\tiny{($\pm$0.005)} & 0.67\tiny{($\pm$0.006)} & 81.7\tiny{($\pm$0.25)} & {82.4}\tiny{($\pm$0.15)} & {80.29}\tiny{($\pm$0.6)} &  {82.8}\tiny{($\pm$±0.08)} & 0.86\tiny{($\pm$0.001)} & 0.86\tiny{($\pm$0.001)}  \\
\midrule
Supervised & & 0.79\tiny{($\pm$0.001)} & & & & & & 0.94\tiny{($\pm$0.001)} \\
\bottomrule
\end{tabular}}
\end{center}
}
\end{table*}

To assess the performance of the models in generating high-quality samples across different modalities, we employ two quantitative evaluation metrics. For the image modality, we utilize the Fréchet Inception Distance (FID) score. On the other hand, for the mask and attribute modalities, we employ the sample-average F1 score as the evaluation metric. The F1 score provides a comprehensive measure of the model's ability to accurately generate binary predictions in multi-label classifications.

We present our results in table ~\ref{celeba_result} which shows the performance of the models in both conditional and unconditional settings. Diff-MVAE generates high-quality images, achieving the best FID scores among all models in image generation. Not only that but also the performance of the model in the other modalities where Diff-MVAE outperforms the baselines in attribute prediction, even if the attribute modality architectures are the same for all models. Our models also have very close results to supervised prediction models that are trained to predict the attribute or mask directly from the image modality ~\citep{wesego}. Unconditional results also show that the baselines do not generate good-quality images unconditionally. The effect of the auxiliary score-based model can be seen in this case, where Diff-MVAE can leverage it to generate coherent outputs along all the modalities. In general, the results indicate that our proposed Diff-MVAE model exhibits superior performance in both conditional and unconditional generation tasks, consistently producing higher-quality and more coherent outputs compared to the baselines. We show qualitative results in Figure ~\ref{fig:mddvae_cond}. The figure shows conditionally generated images for Diff-MVAE and the baselines where Diff-MVAE generates higher-quality images that are more coherent with the given conditions.

\subsection{Ablation: Effect of the Auxiliary Score Model}

\begin{figure}[!h]
    \centering
    \begin{subfigure}[!]{0.6\textwidth}
    \begin{subfigure}[!]{0.48\textwidth}
        \centering
         \includegraphics[width=1\textwidth]{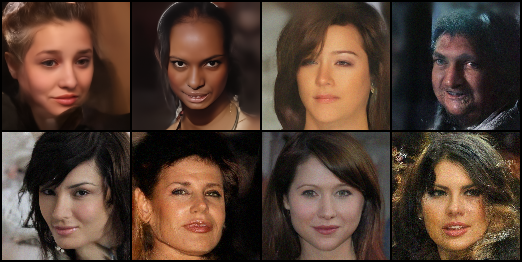}
    \end{subfigure}
    \begin{subfigure}[!]{0.48\textwidth}
        \centering
         \includegraphics[width=1\textwidth]{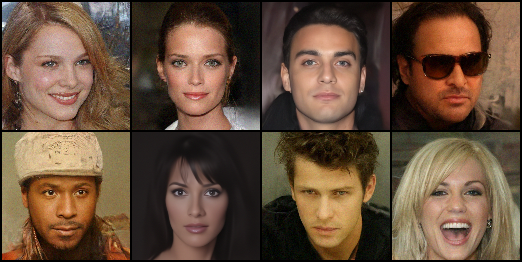}
    \end{subfigure}
    \end{subfigure}
    
    \begin{subfigure}[!]{0.6\textwidth}
    \begin{subfigure}[!]{0.48\textwidth}
        \centering
         \includegraphics[width=1\textwidth]{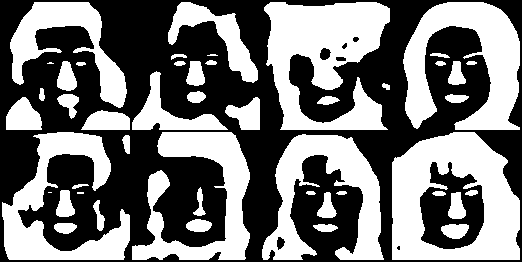}
    \end{subfigure}
    \begin{subfigure}[!]{0.48\textwidth}
        \centering
         \includegraphics[width=1\textwidth]{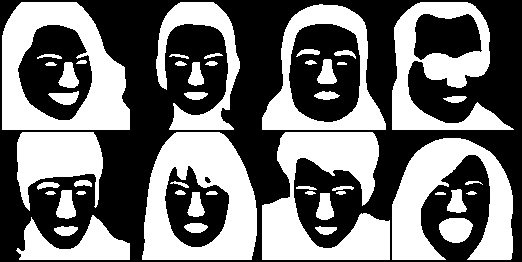}
    \end{subfigure}
    \end{subfigure}
    
    \begin{subfigure}[!]{0.6\textwidth}
    \begin{subfigure}[!]{0.48\textwidth}
    \centering
     \includegraphics[width=1\textwidth]{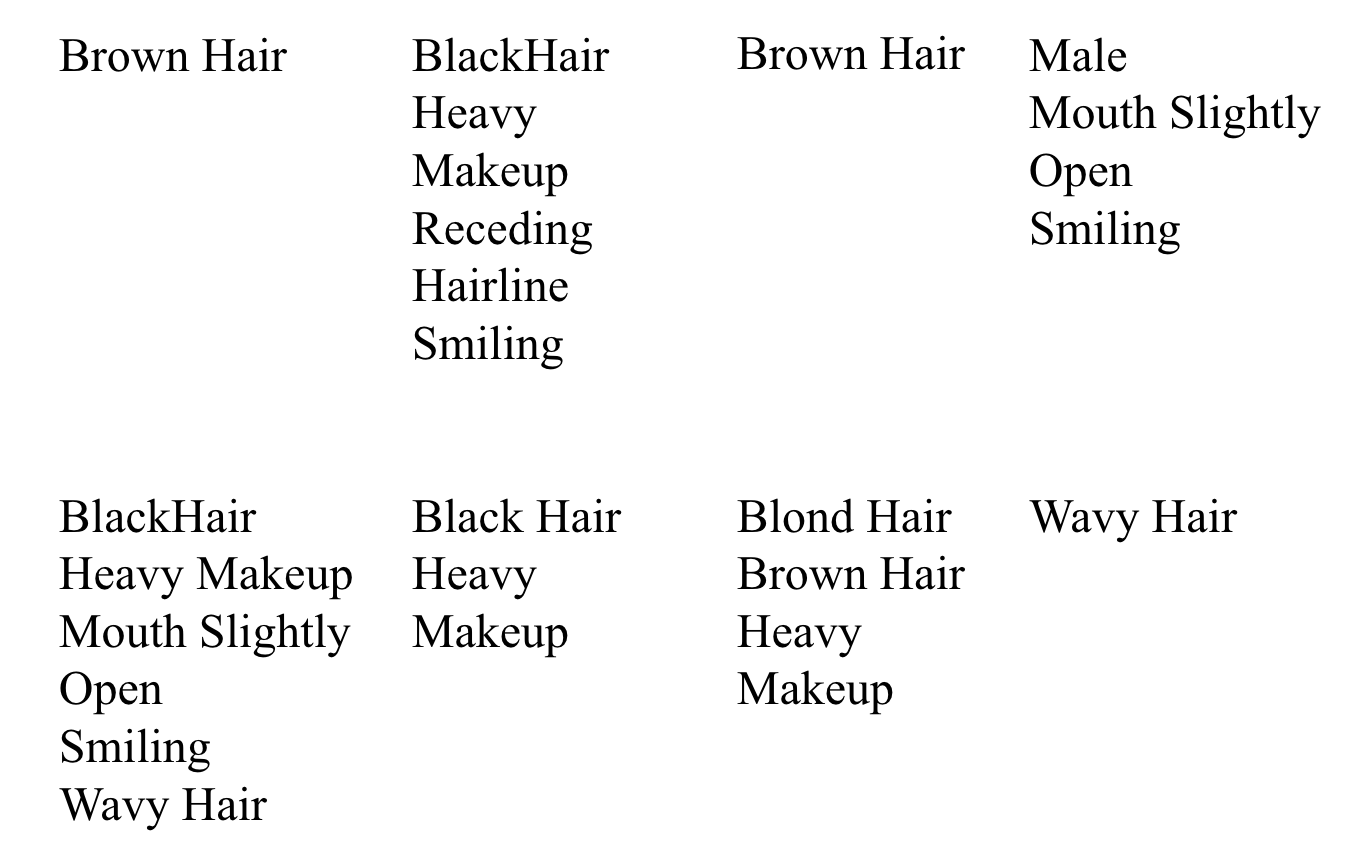}
    \end{subfigure}
    \begin{subfigure}[!]{0.48\textwidth}
    \centering
     \includegraphics[width=1\textwidth]{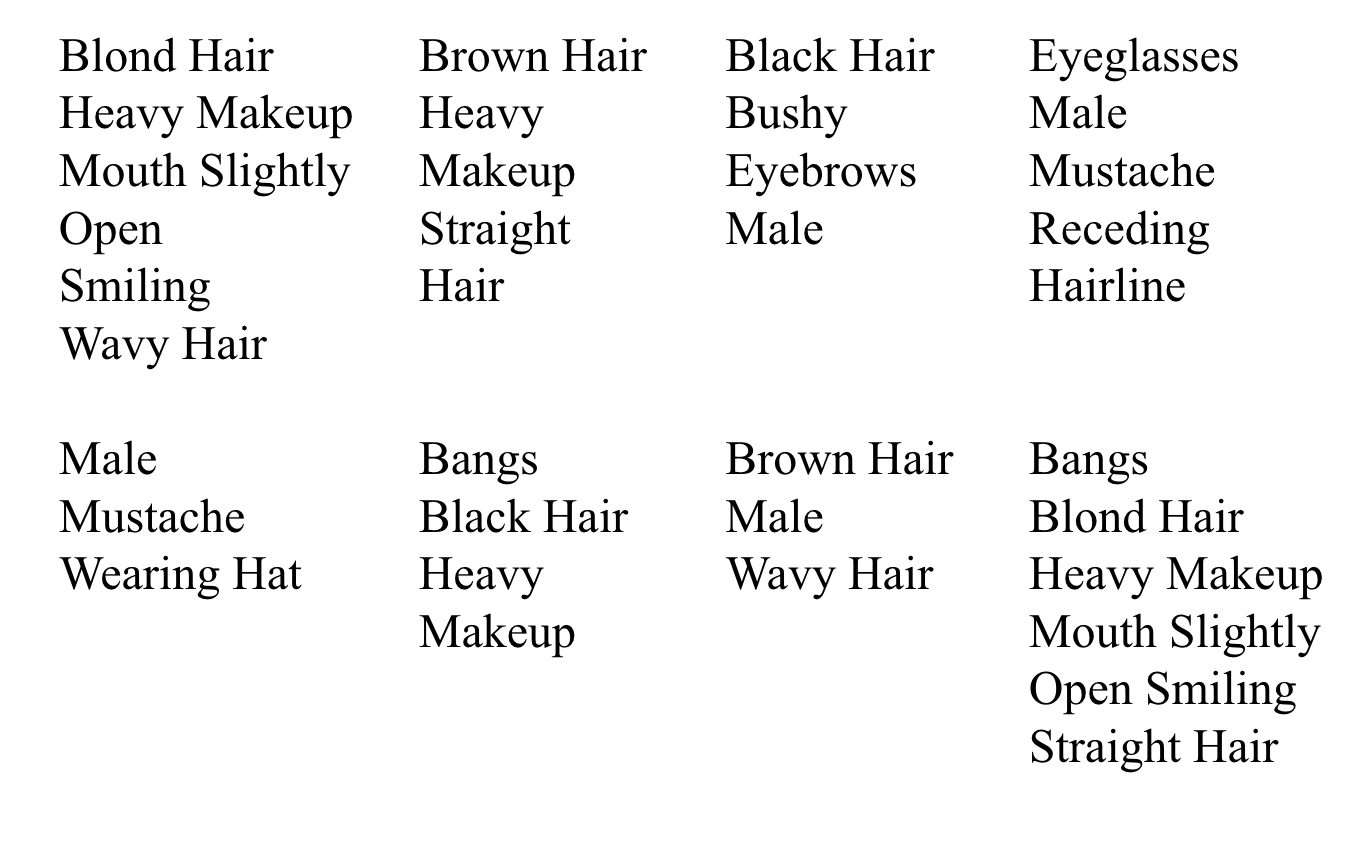}
    \end{subfigure}
    \end{subfigure}
    
    \caption{Unconditional generation using Diff-MVAE* when using no auxiliary score model (left) and when using auxiliary score model (right)}
    \label{fig:mddvae_aux_effect}
\end{figure}

This section covers the importance of using the auxiliary score-based model. By default, multimodal VAEs sample from $\mathcal{N}(0,\mathbf{I})$ to perform unconditional generation. The problem with this is there could be considerable differences in distribution between the prior and the posterior, which will lead to low-quality samples. Previous unimodal VAE works improved sample quality by training an energy-based model on the latent space that will serve as a prior instead of the standard normal distribution and provide samples closer to the posterior, thereby improving generated outputs \citep{ncpvae}. ~\citet{preechakul2021diffusion} also used a score-based model in the unimodal diffusion autoencoder to generate samples unconditionally. In light of these, we train an auxiliary score-based model to generate prior samples for unconditional multi-modal generation. The auxiliary score-based model will transform the standard normal sample to a sample from the posterior $q_\phi (\mathbf{z}|\mathbf{x})$. Figure ~\ref{fig:mddvae_aux_effect} compares outputs with and without the auxiliary score model, showing inferior results on the left and improved ones on the right.

\section{Conclusion and Discussion}
In this paper, we introduce Diff-MVAE, a new multimodal VAE that enhances the capabilities of traditional multimodal VAEs by incorporating diffusion decoders, leading to superior performance. Additionally, we implement an auxiliary model to bridge the gap between the prior and posterior distributions, achieving high-quality and coherent results. A notable limitation of Diff-MVAE is its higher computational demand during sampling due to the iterative diffusion process. Nonetheless, our model demonstrates significant performance improvements across multiple challenging datasets, advancing the field of multimodal VAEs and positioning them as a preferred choice for multimodal applications. Future work will focus on optimizing the multimodal model to reduce computational overhead. Our contributions provide a robust framework for future research and development in multimodal VAE generative modeling.

\bibliography{bibliography}
\bibliographystyle{plainnat}

\clearpage
\section{Appendix}

\subsection{Multimodal ELBO with Diffusion Decoder}
\label{appx:elbo_proof}

The marginal likelihood of the data $ p(\mathbf{x}) $ is given by:
\begin{align}
    \log p(\mathbf{x}) = \log \int p(\mathbf{x}, \mathbf{z}) \, d\mathbf{z},
\end{align}
where $ \mathbf{z} $ is the shared latent variable. Using the definition of joint probability:
\begin{align}
    p(\mathbf{x}, \mathbf{z}) = p(\mathbf{x} | \mathbf{z}) p(\mathbf{z}),
\end{align}
the marginal likelihood becomes:
\begin{align}
    \log p(\mathbf{x}) = \log \int p(\mathbf{x} | \mathbf{z}) p(\mathbf{z}) \, d\mathbf{z}.
\end{align}

To make this integral tractable, we introduce a variational distribution $ q(\mathbf{z} | \mathbf{x}) $, which approximates the true posterior $ p(\mathbf{z} | \mathbf{x}) $. Using Jensen's inequality:
\begin{align}
    \log p(\mathbf{x}) \geq \mathbb{E}_{q(\mathbf{z} | \mathbf{x})} \Big[ \log p(\mathbf{x} | \mathbf{z}) \Big] - \mathbb{D}_{\text{KL}} \big( q(\mathbf{z} | \mathbf{x}) \| p(\mathbf{z}) \big).
\end{align}

This inequality defines the Evidence Lower Bound (ELBO):
\begin{align}
    \text{ELBO} = \mathbb{E}_{q(\mathbf{z} | \mathbf{x})} \Big[ \log p(\mathbf{x} | \mathbf{z}) \Big] - \mathbb{D}_{\text{KL}} \big( q(\mathbf{z} | \mathbf{x}) \| p(\mathbf{z}) \big).
\end{align}

In the multimodal setting, we assume that each modality $ \mathbf{x}_i $ is conditionally independent given $ \mathbf{z} $. Thus:
\begin{align}
    p(\mathbf{x}_M | \mathbf{z}) = \prod_{i \in M} p(\mathbf{x}_i | \mathbf{z}),
\end{align}
where $ i \subseteq \{1, 2, \dots, M\} $ represents the set of observed modalities. Substituting this into the ELBO:
\begin{align}
    \text{ELBO} = \mathbb{E}_{q(\mathbf{z} | \mathbf{x})} \Big[ \sum_{i \in M} \log p(\mathbf{x}_i | \mathbf{z}) \Big] - \mathbb{D}_{\text{KL}} \big( q(\mathbf{z} | \mathbf{x}) \| p(\mathbf{z}) \big).
\end{align}

Now, consider the mixture-based encoder $ q(\mathbf{z} | \mathbf{x}) $ defined as:
\begin{align}
    q(\mathbf{z} | \mathbf{x}) = \sum_{A \in \mathcal{S}} \omega_A q(\mathbf{z} | \mathbf{x}_A),
\end{align}
where $ \mathcal{S} $ is the set of modalities, and $ \omega_A $ is the weighting factor for each subset $ A $ and $\leq 1$.  The ELBO becomes:
\begin{align}
    \text{ELBO} = \mathbb{E}_{\sum_{A \in \mathcal{S}} \omega_A q(\mathbf{z} | \mathbf{x}_A)} \Big[ \sum_{i \in M} \log p(\mathbf{x}_i | \mathbf{z}) \Big] - \mathbb{E}_{\sum_{A \in \mathcal{S}} \omega_A q(\mathbf{z} | \mathbf{x}_A)} \Big[ \log ( q(\mathbf{z} | \mathbf{x})) - \log (p(\mathbf{z})) \Big] 
\end{align}

Taking the sum outside the expectation and separating the second term:
\begin{align}
    = \sum_{A \in \mathcal{S}} \omega_A \mathbb{E}_{q(\mathbf{z} | \mathbf{x}_A)} \Big[ \sum_{i \in M} \log p(\mathbf{x}_i | \mathbf{z}) \Big] - \sum_{A \in \mathcal{S}} \omega_A \mathbb{E}_{q(\mathbf{z} | \mathbf{x}_A)} \log ( q(\mathbf{z} | \mathbf{x})) + \sum_{A \in \mathcal{S}} \omega_A \mathbb{E}_{q(\mathbf{z} | \mathbf{x}_A)} \log (p(\mathbf{z})) 
\end{align}

We can write the second term as:
\begin{align}
    \sum_{A \in \mathcal{S}} \omega_A \mathbb{E}_{q(\mathbf{z} | \mathbf{x}_A)} \log ( q(\mathbf{z} | \mathbf{x})) = \sum_{A \in \mathcal{S}} \omega_A \mathbb{E}_{q(\mathbf{z} | \mathbf{x}_A)} \log ( q(\mathbf{z} | \mathbf{x}_A)) - \sum_{A \in \mathcal{S}} \omega_A D_\text{KL} \big(q_{\phi}(\mathbf{z}|\mathbf{x}_A) \big\| q(\mathbf{z}|\mathbf{x})\big)
\end{align}

Substituting back:
\begin{align*}
     = &  \sum_{A \in \mathcal{S}} \omega_A \mathbb{E}_{q(\mathbf{z} | \mathbf{x}_A)} \Big[ \sum_{i \in M} \log p(\mathbf{x}_i | \mathbf{z}) \Big] + \sum_{A \in \mathcal{S}} \omega_A D_\text{KL} \big(q_{\phi}(\mathbf{z}|\mathbf{x}_A) \big\| q(\mathbf{z}|\mathbf{x})\big)  \\ 
    & - \sum_{A \in \mathcal{S}} \omega_A \mathbb{E}_{q(\mathbf{z} | \mathbf{x}_A)} \log ( q(\mathbf{z} | \mathbf{x}_A)) + \sum_{A \in \mathcal{S}} \omega_A \mathbb{E}_{q(\mathbf{z} | \mathbf{x}_A)} \log (p(\mathbf{z})) 
\end{align*}

Because the KL term is always positive, we can remove it and create a new ELBO:
\begin{align*}
    = &  \sum_{A \in \mathcal{S}} \omega_A \mathbb{E}_{q(\mathbf{z} | \mathbf{x}_A)} \Big[ \sum_{i \in M} \log p(\mathbf{x}_i | \mathbf{z}) \Big] - \sum_{A \in \mathcal{S}} \omega_A \mathbb{E}_{q(\mathbf{z} | \mathbf{x}_A)} \log ( q(\mathbf{z} | \mathbf{x}_A)) + \sum_{A \in \mathcal{S}} \omega_A \mathbb{E}_{q(\mathbf{z} | \mathbf{x}_A)} \log (p(\mathbf{z})) \\
    = &  \sum_{A \in \mathcal{S}} \omega_A \mathbb{E}_{q(\mathbf{z} | \mathbf{x}_A)} \Big[ \sum_{i \in M} \log p(\mathbf{x}_i | \mathbf{z}) \Big] - \sum_{A \in \mathcal{S}} \omega_A D_\text{KL} \big(q_{\phi}(\mathbf{z}|\mathbf{x}_A) \big\| p(\mathbf{z})\big) \\
    = & \sum_{A \in \mathcal{S}} \omega_A \Bigg\{ \mathbb{E}_{q(\mathbf{z} | \mathbf{x}_A)} \Big[ \sum_{i \in M} \log p(\mathbf{x}_i | \mathbf{z}) \Big] -  D_\text{KL} \big(q_{\phi}(\mathbf{z}|\mathbf{x}_A) \big\| p(\mathbf{z})\big) \Bigg\}
\end{align*}

Which gives us the ELBO shown in equation ~\ref{eq:m_elbo_l}:
\begin{align}
    \log &p(\mbfm{x}) \ge \nonumber \sum_{A \in \mathcal{S}} \omega_A \Bigg\{ \mathbb{E}_{q(\mathbf{z} | \mathbf{x}_A)} \Big[ \sum_{i \in M} \log p(\mathbf{x}_i | \mathbf{z}) \Big] -  D_\text{KL} \big(q_{\phi}(\mathbf{z}|\mathbf{x}_A) \big\| p(\mathbf{z})\big) \Bigg\}
\end{align}

\subsubsection{Restating Proposition 1} \label{appx:diff_elbo_proof}
Now, adding the diffusion decoder part, the objective in Equation ~\ref{eq:m_elbo_l3} is given by:
\begin{align}
    \sum_{A \in \mathcal{S}} \omega_A \Bigg\{ 
    \mathbb{E}_{q_\phi(\mathbf{z} | \mathbf{x}_A)} \Bigg[ 
    \sum_{i=1}^M \mathbb{I}_{\text{ff}(i)} \log p_\theta(\mathbf{x}_i | \mathbf{z}) 
    + \mathbb{I}_{\text{diff}(i)} \mathbb{E}_{\mathbf{x}_{it}} \frac{\lambda(t)}{2} 
    \Big\| \nabla_{\mathbf{x}_{it}} \log p_t(\mathbf{x}_{it} | \mathbf{x}_i, \mathbf{z}) \\
    - s_\theta(\mathbf{x}_{it}, \mathbf{z}, t) \Big\|^2 \Bigg] \nonumber - \mathbb{D}_{\text{KL}} \big(q_\phi(\mathbf{z} | \mathbf{x}_A) \| p(\mathbf{z}) \big) \Bigg\}.
\end{align}

where $\mathbb{I}_{\text{diff}^{(i)}}$ as the indicator function for using a diffusion decoder for modality $i$, and $\mathbb{I}_{\text{ff}^{(i)}}$ as the indicator function for using a standard feed-forward decoder for modality $i$

Our aim is to show that this objective is a valid lower bound on the marginal likelihood of the data $ p(\mathbf{x}) $. Since we can only have one decoder for one modality, we only need to show that when using each type of decoder, we are optimizing a lower bound.

For a feed-forward decoder of a single modality, this directly corresponds to the Evidence Lower Bound (ELBO):
\begin{align}
    \text{ELBO}_{\text{ff}} = \mathbb{E}_{q_\phi(\mathbf{z} | \mathbf{x})} \Big[ \log p_\theta(\mathbf{x} | \mathbf{z}) \Big] - \mathbb{D}_{\text{KL}} \big( q_\phi(\mathbf{z} | \mathbf{x}) \| p(\mathbf{z}) \big).
\end{align}

For a diffusion decoder of a single modality, we model $ p_\theta(\mathbf{x} | \mathbf{z}) $ using the score-based framework. Instead of directly optimizing $ \log p_\theta(\mathbf{x} | \mathbf{z}) $, we minimize the score-matching loss:
\begin{align}
    \mathbb{E}_{\mathbf{x}_{it}} \frac{\lambda(t)}{2} \Big\| \nabla_{\mathbf{x}_{it}} \log p_t(\mathbf{x}_{it} | \mathbf{x}, \mathbf{z}) - s_\theta(\mathbf{x}_{it}, \mathbf{z}, t) \Big\|^2.
\end{align}

To train a model to maximize a probability distribution $p(\mathbf{y})$, we can optimize a model with parameters $\theta$ and minimize the KL divergence between $p(\mathbf{y})$ and $p_{\theta}(\mathbf{y})$ which is $D_{\text{KL}}(p(\mathbf{y}) \parallel p_\theta(\mathbf{y}))$. When we use the SDE described in this work, ~\citet{song2021maximum} shows that: $D_{\text{KL}}(p \parallel p_{\theta}^{\text{SDE}}) \leq \mathcal{J}_{\text{SM}}(\theta; g(\cdot)^2) + D_{\text{KL}}(p_T \parallel \pi).$ By design of the forward SDE process, the second term goes to zero and is not dependent on the model parameters. $\mathcal{J}_{\text{SM}}$ is the exact score-matching loss, which is unknown but can be trained by approximating the score function using denoising score-matching loss shown in equation ~\ref{eq:sm_ll} which is also a lower bound as $D_{\text{KL}}(p \parallel p_{\theta}^{\text{SDE}}) \leq \mathcal{J}_{\text{DSM}}(\theta; g(\cdot)^2)$ \citep{song2021maximum}. This loss ensures that the model approximates the gradient of the log-likelihood $\nabla_{\mathbf{x}} \log p(\mathbf{x} | \mathbf{z}) \approx s_\theta(\mathbf{x}, \mathbf{z})$ and indirectly maximizes the likelihood $ p(\mathbf{x} | \mathbf{z})$ as the loss is its lower bound. Thus, the score-matching loss serves as a surrogate for the likelihood term in the ELBO. This can be seen as another way of optimizing $p(\mathbf{x} | \mathbf{z})$ using the denoising score matching loss which will optimize the lower bound on the likelihood, $\mathbb{E}_{q(\mathbf{z} | \mathbf{x})} \log(p(\mathbf{x} | \mathbf{z}) \geq \mathbb{E}_{q(\mathbf{z} | \mathbf{x})} \log(p_\theta^{\text{SDE}}(\mathbf{x} | \mathbf{z})$.
\begin{align}
    & \log p(\mathbf{x}) \geq \mathbb{E}_{q(\mathbf{z} | \mathbf{x})} \Big[ \log p(\mathbf{x} | \mathbf{z}) \Big] - \mathbb{D}_{\text{KL}} \big( q(\mathbf{z} | \mathbf{x}) \| p(\mathbf{z}) \big) \geq \mathbb{E}_{q(\mathbf{z} | \mathbf{x})} \Big[ \log(p_\theta^{\text{SDE}}(\mathbf{x} | \mathbf{z}) \Big] \nonumber \\
    &  - \mathbb{D}_{\text{KL}} \big( q(\mathbf{z} | \mathbf{x}) \| p(\mathbf{z}) \big).  \\
    & \log p(\mathbf{x}) \geq
    \mathbb{E}_{q(\mathbf{z} | \mathbf{x})} \Big[ \log(p_\theta^{\text{SDE}}(\mathbf{x} | \mathbf{z}) \Big] - \mathbb{D}_{\text{KL}} \big( q(\mathbf{z} | \mathbf{x}) \| p(\mathbf{z}) \big).
\end{align}

So, the ELBO for using diffusion decoder will be:
\begin{align}
    \text{ELBO}_{\text{dd}} = \mathbb{E}_{q_\phi(\mathbf{z} | \mathbf{x})} \Big[ \mathbb{E}_{\mathbf{x}_{it}} \frac{\lambda(t)}{2} 
    \Big\| \nabla_{\mathbf{x}_{it}} \log p_t(\mathbf{x}_{it} | \mathbf{x}_i, \mathbf{z}) - s_\theta(\mathbf{x}_{it}, \mathbf{z}, t) \Big\|^2 \Big] - \mathbb{D}_{\text{KL}} \big( q_\phi(\mathbf{z} | \mathbf{x}) \| p(\mathbf{z}) \big).
\end{align}

\subsubsection*{Combining the Objectives}
The combined objective, which will give us Equation ~\ref{eq:m_elbo_l3}, incorporates both feed-forward and diffusion decoders. Only one of the decoders will be used for each modality, and both terms contribute to maximizing the marginal likelihood of the data $ p(\mathbf{x}) $, as the score-matching loss provides a valid surrogate for the likelihood term $ \log p_\theta(\mathbf{x} | \mathbf{z}) $ which makes the overall equation a lower bound on the probability of the data distribution.
\begin{align}
\log &p(\mbfm{x}) \ge \nonumber \sum_{A \in \mathcal{S}} \omega_A \Bigg\{ \mathbb{E}_{q_{\phi}(\mathbf{z}| \mathbf{x}_A)} \Bigg[ \sum_{i=1}^{M_F} \log p_\theta(\mathbf{x}_i| \mathbf{z}) + \nonumber \sum_{j=1}^{M_D} \mathbb{E}_{\mathbf{x}_{jt}} 
 \frac{1}{2} \lambda(t) \bigg\| \nabla_{\mathbf{x}_{jt}} \log p_t(\mathbf{x}_{jt} \mid \mathbf{x}_j, \mathbf{z}) - \nonumber \\ 
 & s_\theta(\mathbf{x}_{jt}, \mathbf{z}, t) \bigg\|_2^2 \Bigg] - D_\text{KL} \big(q_{\phi}(\mathbf{z}|\mathbf{x}_A) \big\| p(\mathbf{z})\big) \Bigg\}.
\end{align}

\subsection{PolyMnist Experiment}

\begin{figure}
    \label{polymnist_exp}
    \centering
    \begin{subfigure}{0.32\linewidth}
        \centering
        \includegraphics[width=\linewidth]{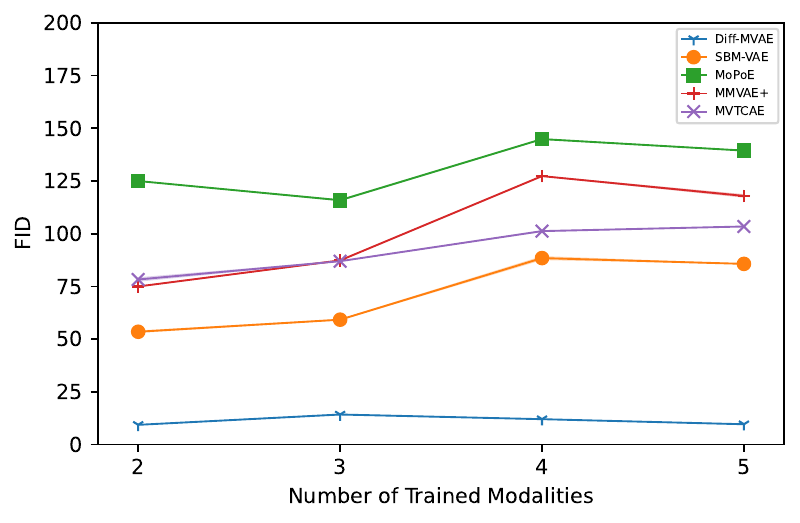}
        \caption{Unconditional FID}
        \label{fig:unc_fid_poly}
    \end{subfigure}
    \begin{subfigure}{0.32\linewidth}
        \centering
        \includegraphics[width=\linewidth]{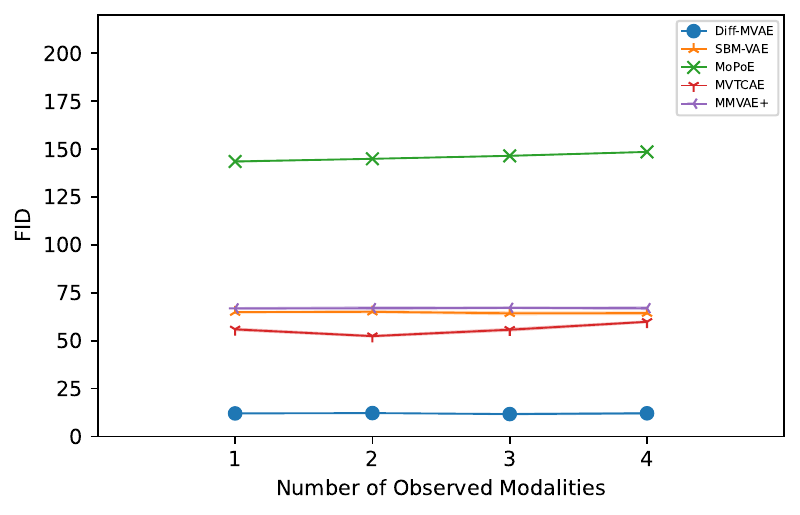}
        \caption{Conditional FID 2}
        \label{fig:cond_fid_poly}
    \end{subfigure}
    \begin{subfigure}{0.32\linewidth}
        \centering
        \includegraphics[width=\linewidth]{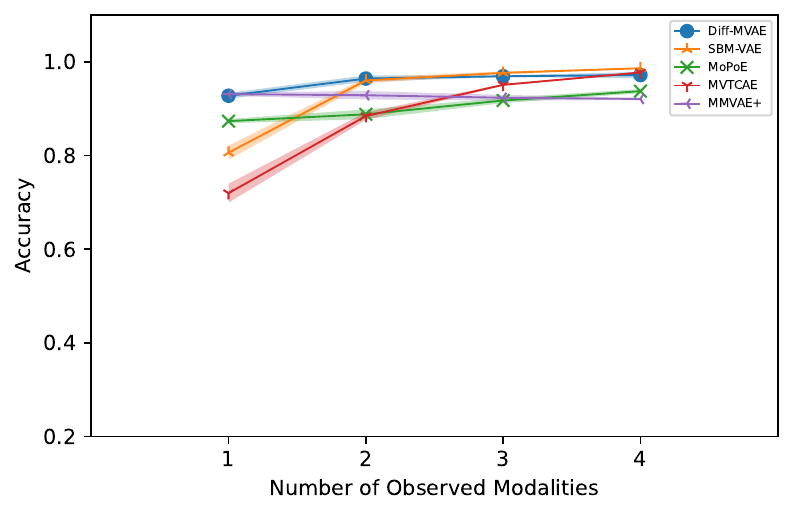}
        \caption{Conditional Accuracy}
        \label{fig:cond_acc_poly}
    \end{subfigure}
    \caption{Performance on 5 modalities of Ext-PolyMnist dataset}
    \label{fig:three_subfigures}
\end{figure}

In this section, we add 5 modalities from the Extended PolyMnist dataset \citep{wesego}. This dataset is mainly used in many of the baselines because it is easier to study the properties of the model under different number of modalities without being affected by computation power. We show the performance of the models in unconditional and conditional settings. First, figure ~\ref{fig:unc_fid_poly} shows the unconditional performance of the models when the number of modalities the model is trained is increased from 2 to 5 modalities. ~\citet{daunhawer2022on} showed that mixture-based multimodal VAEs experience performance degradation as the number of modalities the model is trained on increases. Diff-MVAE* shows almost a horizontal line, showing that the model doesn't lose performance even when the modalities are high. Figure ~\ref{fig:cond_fid_poly} and ~\ref{fig:cond_acc_poly} show models trained on 5 modalities and how the models perform conditionally. The number of observed modalities to generate the last modality is slowly increased and FID and the accuracy are calculated. As shown in the figures, Diff-MVAE* has superior FID and very good accuracy. Overall, our proposed model has better conditional coherence while having the best image generation quality. 

\subsection{CUB experiment setup}
\label{appx:cub_setup}

As discussed in the main paper, the CUB dataset consists of two modalities: image and text. The text VAE was initialized from the pre-trained weight of ~\citet{li2020_Optimus} that uses a BERT encoder and GPT-2 decoder for all models. The images are resized to 64x64. We used a latent size of $768$. The encoder and decoder architecture for the image modality are almost identical to that of ~\citet{softvae} except for the diffusion decoder, which uses a UNET architecture. The architecture with all the details can be referred in the attached code. The models were trained for 500 epochs using the Adam optimizer \citep{adam}. The baseline models are trained using different $\beta$ values from 0.1, 0.5, 1, and 5, and the best model was selected by using the average of the conditional and unconditional FID. For MoPoE, $\beta$ value of $5.0$ is selected, while for the MVTCAE, $\beta$ value of $1.0$ is selected. We used a $\lambda$ of $1e-5$ for Diff-MVAE variants.

The unconditional auxiliary score model also uses a 1D UNET architecture and accepts input similar to the size of the latent dimension ($768$). We use DDIM sampling for both the Diff-MVAE* and the auxiliary score-based model with 50 sampling steps. For Diff-MVAE, we use Euler-Maruyama sampling for 1000 steps. For training, we use continuous score-based diffusion for Diff-MVAE, and discrete timesteps for Diff-MVAE* similar to most DDPM training setup. For calculating FID and Clip-Score, 10000 samples were used from the test set. The main Diff-MVAE variants are trained on 2 A100 GPUs. The training time it takes is approximately about 150 hours. The auxiliary model is trained on 1 A100 GPU for approximately about 40 hours.

\subsection{CelebAMask-HQ experiment setup}
\label{appx:celeba_setup}
The CelebMaskHQ dataset is taken from \citet{CelebAMask-HQ} where the three modalities are images, masks, and attributes. All face part masks were combined into a single black-and-white image except the skin mask. Out of the 40 attributes, 18 were taken from it similar to the setup of \citep{poe}. The encoder and decoder architectures are similar to \citet{softvae} except MMVAE+ which uses an image encoder and decoder similar to the one presented in their work. A latent size of 256 was used for Diff-MVAE and MMVAE+, and a latent size of 1024 was used for MVTCAE and MoPoE. For MMVAE+, modality-specific and shared latent sizes are each 128 trained with IWAE estimator with K=1. We select the best $\beta$ for the baselines from [0.1,0.5,1,2.5,5]. MoPoE use $\beta$ of 1e-5 with the loss averaged over the dimensions of a data as that gave better result and to make comparison similar to Diff-MVAE. MVTCAE 0.5 and MMVAE+ 5. The diffusion decoder and the auxiliary score-based model use a UNET architecture. The architecture with all the details can be referred to in the attached code. The training and sampling of the Diff-MVAE variants are similar to the ones used in the CUB dataset ~\ref{appx:cub_setup}. For calculating FID and F1-Score, the test set samples were used. The main Diff-MVAE model is trained on 4 A100 GPUs. The training time it takes is approximately about 120 hours. The auxiliary model is trained on 1 A100 GPU for approximately about 2 to 3 hours.

\subsection{More Ablation}

In this section, we study the effect of using the auxiliary score-based model on the baselines. Specifically, we select the MoPoE baseline and train a score-based diffusion prior on the product of experts of the posterior. After training the score-based prior, we generate unconditional $z$ from the trained prior instead of a standard normal distribution to perform unconditional generation. We show the result in Table ~\ref{tab:unc_prior_mopoe}. The FID score is much better when using the unconditional auxiliary score-based prior and the approach can be applied to not only our models but also to other baselines.

\begin{table}[!h]
\caption{Auxiliary Score Model Prior for MoPoE}
\label{tab:unc_prior_mopoe}
\begin{center}
\begin{tabular}{ccc}
\toprule
{} & {Unc (Gaussion Prior)} & {Unc (Score Prior)} \\
\midrule
MoPoE  & 139.8 &  95.2\\

\bottomrule
\end{tabular}
\end{center}
\end{table}

\subsection{Additional Samples}
\begin{figure*}[!h]
    \centering
    \begin{subfigure}[!]{1\textwidth}
    \begin{subfigure}[!]{0.31\textwidth}
        \centering
         \includegraphics[width=1\textwidth]{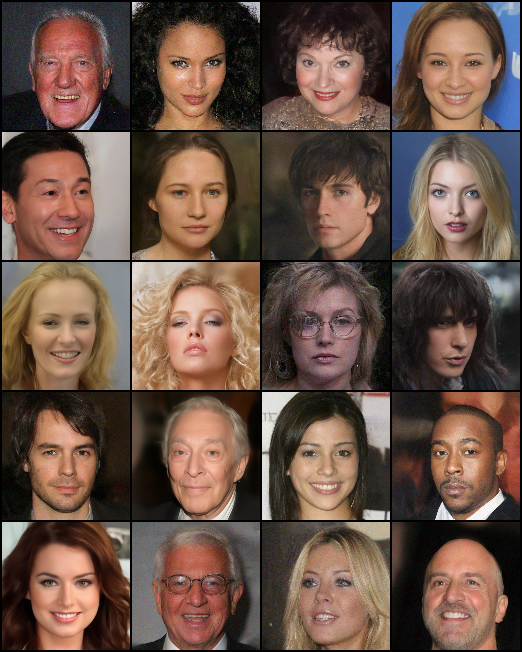}
         \caption{Diff-MVAE* unc image}
    \end{subfigure}
    \begin{subfigure}[!]{0.31\textwidth}
        \centering
         \includegraphics[width=1\textwidth]{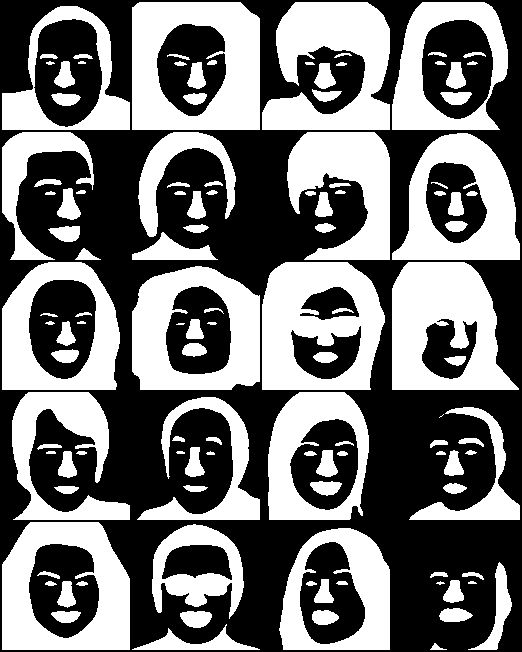}
         \caption{Diff-MVAE* unc mask}
    \end{subfigure}
    \begin{subfigure}[!]{0.4\textwidth}
    \centering
     \includegraphics[width=1\textwidth]{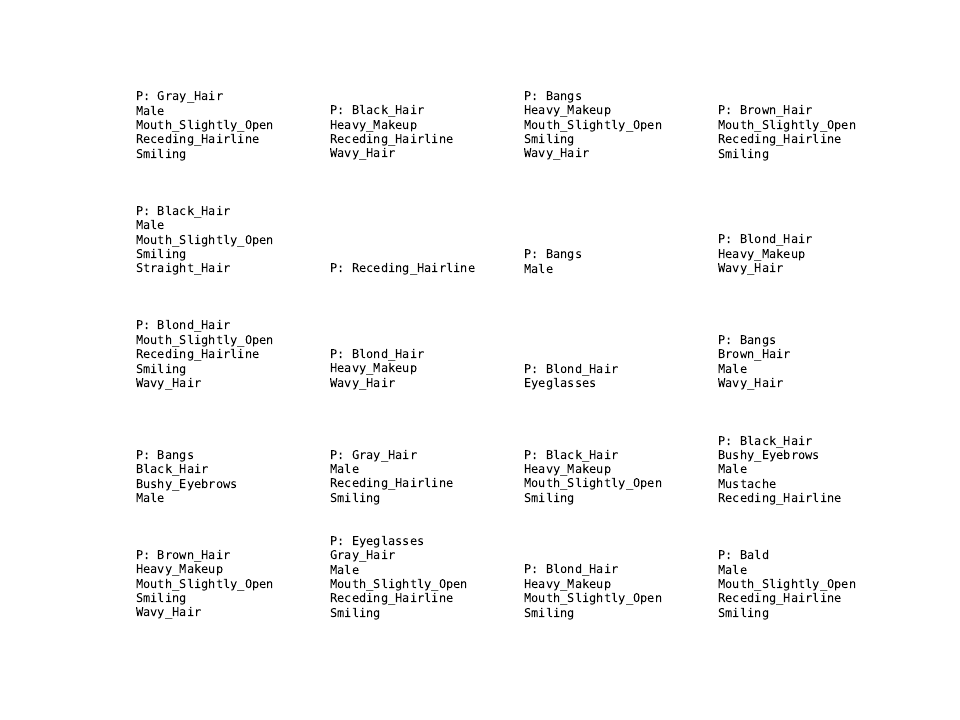}
     \caption{Diff-MVAE* unc attribute}
    \end{subfigure}
    \end{subfigure}
    \caption{Unconditional generation using Diff-MVAE}
    \label{fig:mddvae_unc_appx}
\end{figure*}


\begin{figure*}[!h]
    \centering
    \begin{subfigure}[!]{1\textwidth}
    \begin{subfigure}[!]{0.38\textwidth}
    \centering
     \includegraphics[width=1\textwidth]{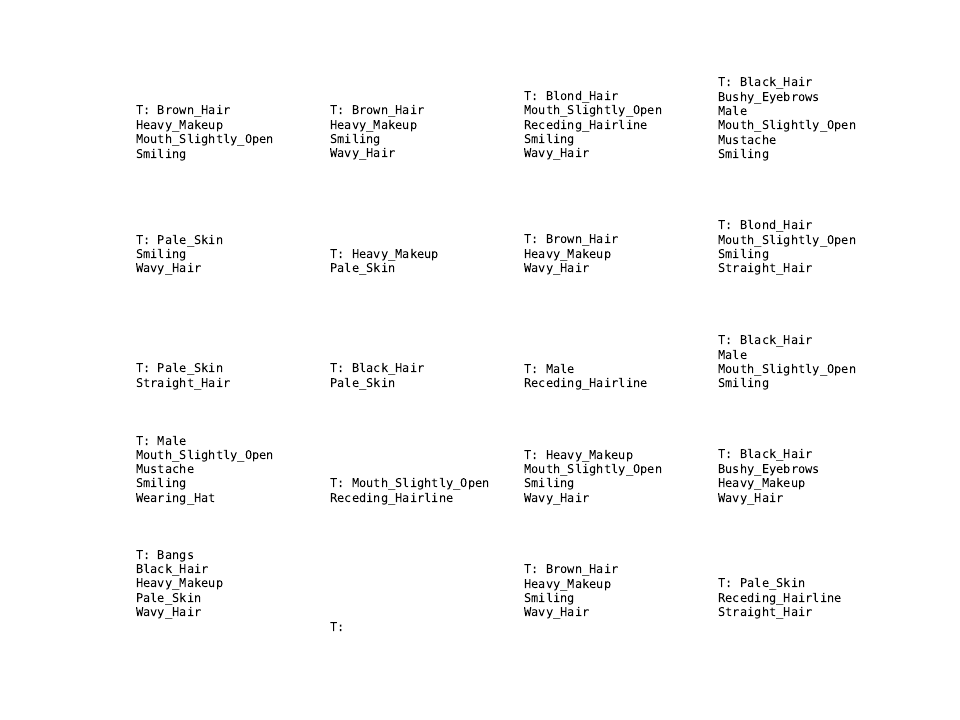}
     \caption{Input attribute}
    \end{subfigure}
    \begin{subfigure}[!]{0.3\textwidth}
        \centering
         \includegraphics[width=1\textwidth]{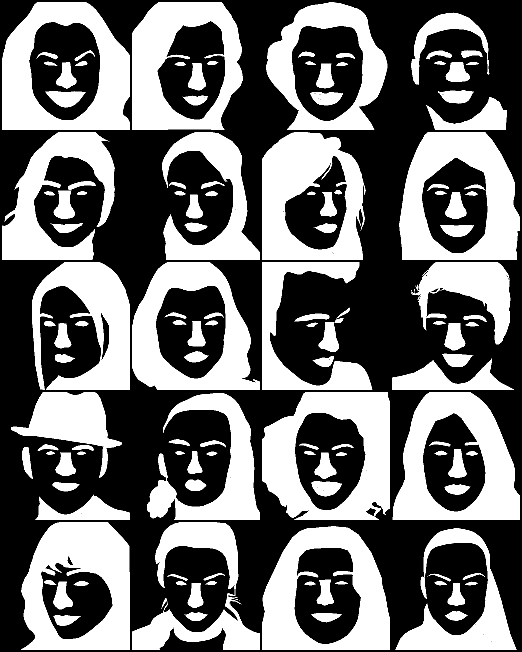}
         \caption{Input mask}
    \end{subfigure}
    \begin{subfigure}[!]{0.3\textwidth}
        \centering
         \includegraphics[width=1\textwidth]{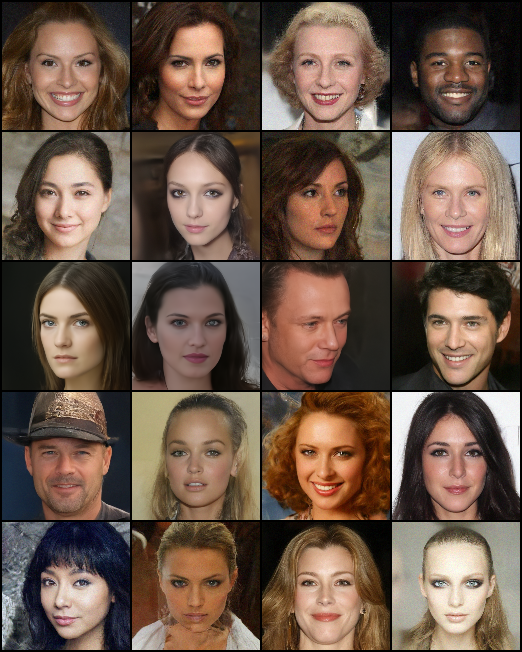}
         \caption{Diff-MVAE* conditional image}
    \end{subfigure}
    \end{subfigure}
    \caption{Conditional generation using Diff-MVAE* given mask and attribute}
    \label{fig:mddvae_given_mask_attr_appx}
\end{figure*}

\begin{figure*}[!h]
    \centering
    \begin{subfigure}[!]{1\textwidth}
    \begin{subfigure}[!]{0.5\textwidth}
    \centering
     \includegraphics[width=1\textwidth]{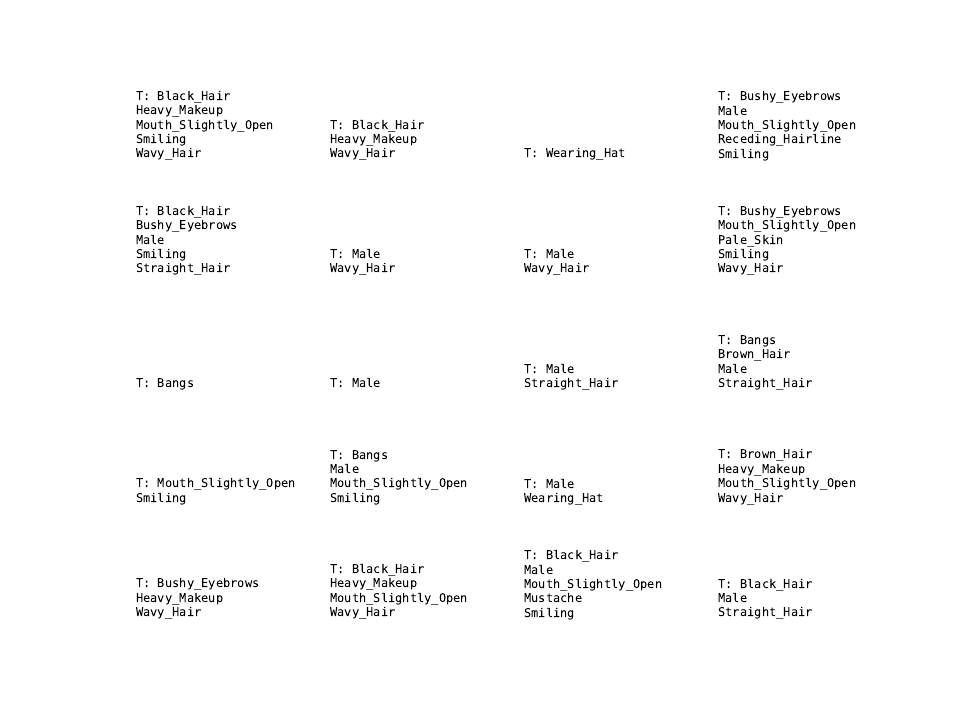}
     \caption{Input attribute}
    \end{subfigure}
    \begin{subfigure}[!]{0.4\textwidth}
        \centering
         \includegraphics[width=1\textwidth]{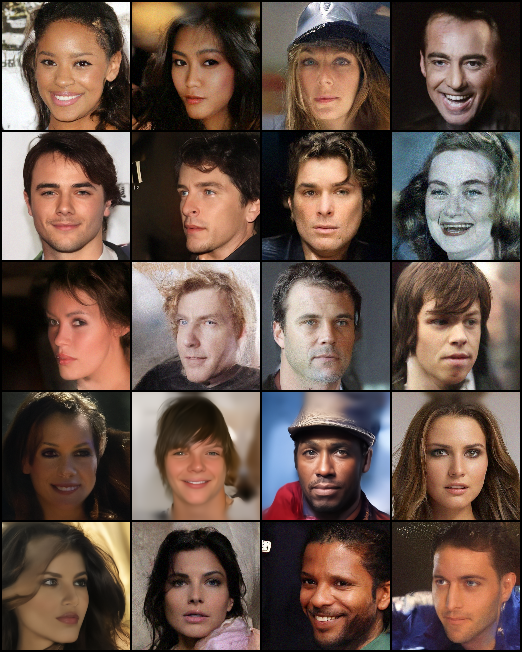}
         \caption{Diff-MVAE* given attribute}
    \end{subfigure}
    \end{subfigure}
    \caption{Conditional generation using Diff-MVAE* given attribute}
    \label{fig:mddvae_given_attr_appx}
\end{figure*}

\begin{figure*}[!h]
    \centering
    \begin{subfigure}[!]{1\textwidth}
    \centering
    \begin{subfigure}[!]{0.4\textwidth}
    \centering
     \includegraphics[width=1\textwidth]{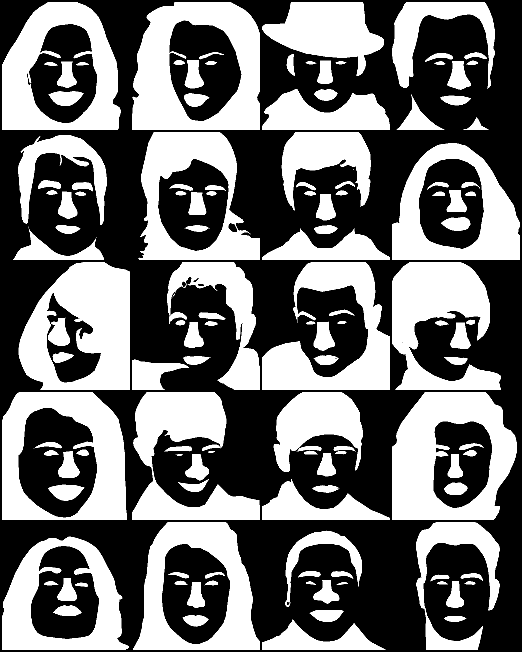}
     \caption{Input mask}
    \end{subfigure}
    \begin{subfigure}[!]{0.4\textwidth}
        \centering
         \includegraphics[width=1\textwidth]{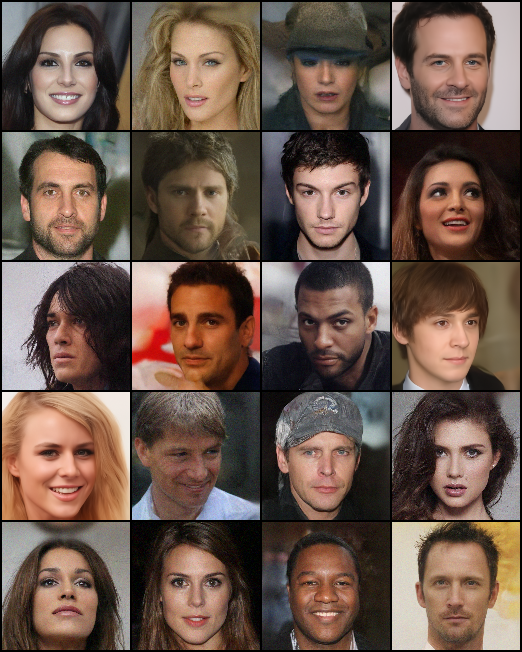}
         \caption{Diff-MVAE* given mask}
    \end{subfigure}
    \end{subfigure}
    \caption{Conditional generation using Diff-MVAE* given mask}
    \label{fig:mddvae_given_mask_appx}
\end{figure*}

\end{document}